\pdfoutput=1

\documentclass[11pt]{article}

\usepackage[final]{acl}
\usepackage{xurl}
\usepackage{longtable}
\usepackage{times}
\usepackage{latexsym}
\usepackage{algorithm}
\usepackage{algpseudocode}
\usepackage{tcolorbox}
\usepackage{multirow}
\usepackage{booktabs}
\usepackage{makecell}
\usepackage[T1]{fontenc}

\usepackage[utf8]{inputenc}
\usepackage{amsfonts}

\usepackage{microtype}

\usepackage{inconsolata}

\usepackage{graphicx}

\usepackage{enumitem}

\usepackage{amsmath}
\usepackage{multicol}
\usepackage[normalem]{ulem}

\setlist[itemize]{itemsep=4pt, topsep=5pt, partopsep=0pt, parsep=0pt}
\setlist[enumerate]{itemsep=4pt, topsep=5pt, partopsep=0pt, parsep=0pt}
\usepackage{subcaption}
\usepackage{caption}
\usepackage{adjustbox}
\usepackage{soul}

\usepackage{tabularx}
\usepackage{array}
\newcolumntype{C}[1]{>{\centering\arraybackslash}p{#1}}

\title{Stress-Testing Emotional Support Models:\\Moving from Homogeneous to Diverse Help Seekers} 

\author{
Chaewon Heo\quad
Cheyon Jin\quad
Yohan Jo\thanks{Corresponding author}\
\\  
Graduate School of Data Science, Seoul National University
\\
\texttt{\{heorshey99, cheyonjin, yohan.jo\}@snu.ac.kr}
}

\begin{document}
\maketitle
\begin{abstract} 
    As emotional support chatbots have recently gained significant traction across both research and industry, a common evaluation strategy has emerged: use help-seeker simulators to interact with supporter chatbots.
    However, current simulators suffer from two critical limitations: (1) they fail to capture the behavioral diversity of real-world seekers, often portraying them as overly cooperative, and (2) they lack the controllability required to simulate specific seeker profiles. To address these challenges, we present a controllable seeker simulator driven by nine psychological and linguistic features that underpin seeker behavior. Using authentic Reddit conversations, we train our model via a \textbf{Mixture-of-Experts (MoE)} architecture, which effectively differentiates diverse seeker behaviors into specialized parameter subspaces, thereby enhancing fine-grained controllability. Our simulator achieves superior profile adherence and behavioral diversity compared to existing approaches. Furthermore, evaluating 7 prominent supporter models with our system uncovers previously obscured performance degradations. These findings underscore the utility of our framework in providing a more faithful and stress-tested evaluation for emotional support chatbots.\footnote{Our code and data are available at \url{https://github.com/holi-lab/ES-Evaluation-Framework}.}

\end{abstract}

\section{Introduction}\label{sec:introduction}
Emotional support is essential for alleviating psychological distress and maintaining mental well-being. While LLM-based emotional support systems have advanced rapidly \cite{Zheng_2025}, the lack of reliable automated evaluation frameworks remains a critical bottleneck, hindering the field’s establishment as an objective and measurable discipline. 

Currently, the most prevalent evaluation strategy uses \textbf{help-seeker simulators} to interact with supporter models, and the resulting dialogues are assessed based on metrics such as empathy and fluency \cite{zhao-etal-2024-esc}.
However, existing seeker simulators compromise evaluation validity due to two critical limitations. First, they fail to reflect the diversity of real-world seekers, representing only a narrow subset of seeker behaviors. Existing simulators primarily portray overly cooperative, "easy" seekers, failing to capture the vast spectrum of real-world behaviors such as advice resistance \cite{Yaman2021Resistance} or limited disclosure. Second, existing simulators lack fine-grained controllability. A robust evaluation framework must allow researchers to target specific seeker populations, as this enables the identification of which aspects of a supporter model fail for particular seeker groups, providing actionable signals for model improvement.


\begin{figure*}[t]
  \hspace*{-5mm}
  \centering
  \includegraphics[width=1.02\textwidth]{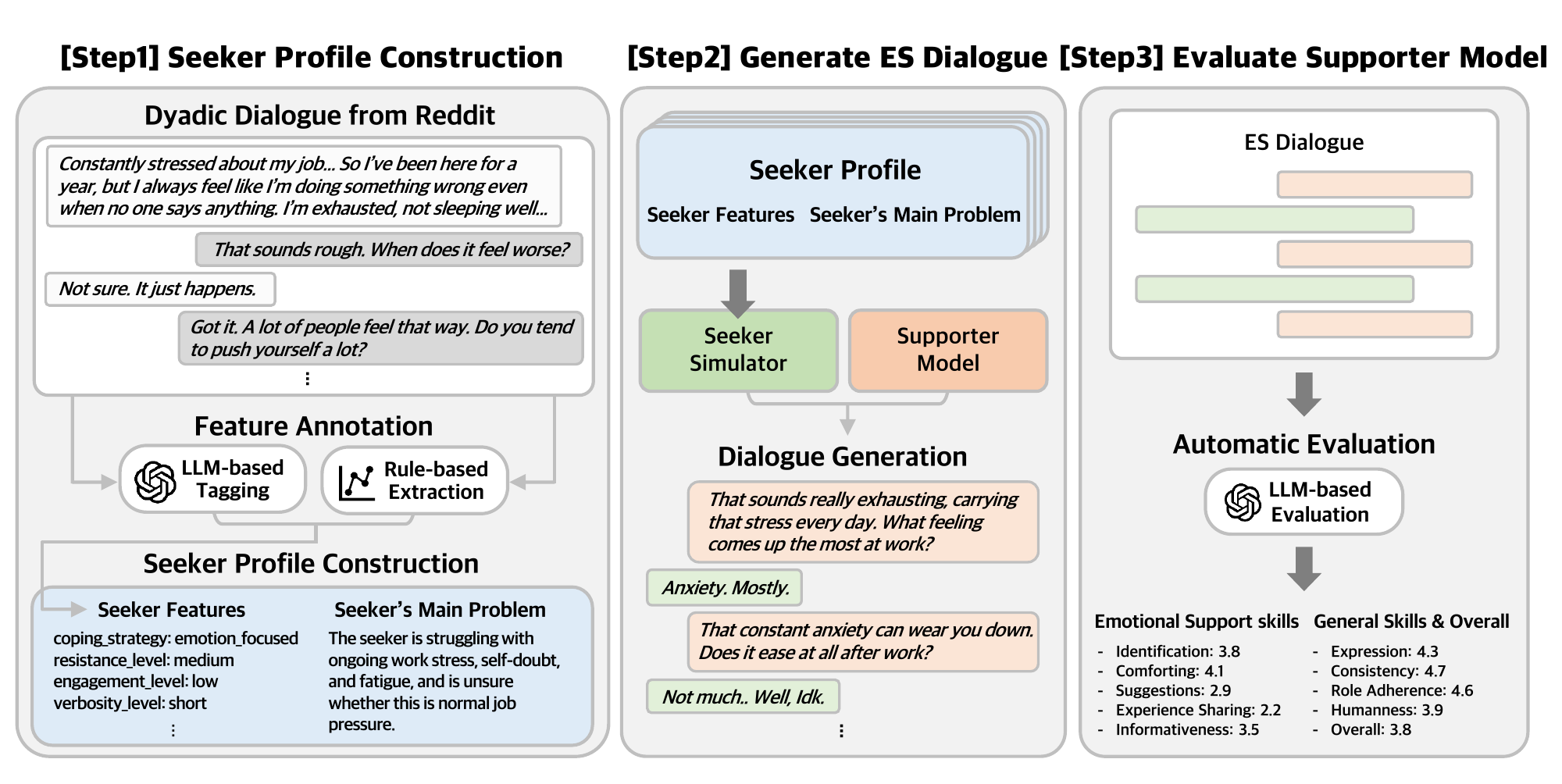}
  \caption{Overview of our evaluation framework for emotional support models.}
  \label{fig:1-overview}
\end{figure*}

To address these shortcomings, we present a highly controllable seeker simulator designed to take specific seeker profiles and simulate their corresponding behaviors. As a foundational component, we define the \textbf{seeker profile} as a combination of nine psychological and linguistic features such as resistance level and verbosity level (Step 1 in Figure~\ref{fig:1-overview}). To faithfully reflect a broad array of real-world behaviors, we leverage data from Reddit online support groups as the primary training source for our simulator.



To effectively control a seeker profile defined by nine features, we employ a Mixture-of-Experts (MoE) architecture. This architecture dynamically assigns weights to specialized adapters based on the input seeker profile, accurately reflecting distinct behavioral patterns across seeker types.

Validation results highlight the effectiveness of our simulator across three key dimensions: 
(1) \textbf{superior profile adherence}, successfully maintaining consistent seeker traits throughout the interaction; 
(2) \textbf{high behavioral diversity}, capturing a wider array of help-seeking patterns compared to existing baselines; and 
(3) \textbf{higher fidelity}, as confirmed by expert evaluations indicating that the generated responses more closely resemble real-world emotional support interactions.

Using our seeker simulator, we evaluate supporter models through interactive multi-turn dialogues, followed by automatic assessment using emotional support and general conversational skill metrics (Steps 2, 3 in Figure~\ref{fig:1-overview}). Results indicate that supporter models exhibit varying levels of performance degradation across various seeker populations. Our framework allows researchers to flexibly define seeker populations by selecting appropriate combinations of profile features, enabling targeted and diagnostic evaluation under diverse scenarios.

In summary, we introduce a controllable seeker simulator driven by multi-dimensional seeker profiles, providing a population-diverse evaluation framework for assessing supporter models.





\section{Related Works}\label{sec:related_works}
\subsection{Evaluation of Emotional Support Models}
The evaluation of emotional support models has evolved from surface-level fluency to interactional supportive efficacy. Early research evaluated whether generated empathetic responses exhibited emotionally appropriate expressions \cite{santhanam-shaikh-2019-emotional}, using lexical metrics (e.g., BLEU, Perplexity) against reference dialogues such as EmpatheticDialogues \cite {rashkin-etal-2019-towards}. Since these metrics often fail to capture supportive effectiveness, later work adopted ESConv \cite{liu-etal-2021-towards} as evaluation scenarios and performed human evaluation, where evaluators role-play seekers and rate support quality using predefined criteria (e.g., Identification, Empathy) \cite{zhao-etal-2023-transesc, cheng-etal-2022-improving}. However, these human role-play evaluations are constrained by predefined scenarios, fail to reflect real-world interaction dynamics, and are costly to scale. These limitations have driven recent evaluation toward simulator-based evaluation frameworks, where seeker simulators interact with supporter models and automated evaluation is performed on the generated dialogues.

\subsection{Seeker Simulations in Emotional Support}
Prior studies have adopted diverse approaches to implement seeker simulators. Prompt-based simulators are given input including specific characteristics, such as Big Five personality traits, or situational mental health contexts \cite{madani-srihari-2025-esc, qiu2024interactiveagentssimulatingcounselorclient}.
Beyond simple prompting, some work employs explicit state tracking or memory modules to regulate the seeker’s internal states \cite{yang-etal-2025-consistent, wang-etal-2025-annaagent}. Another work trains simulators on emotional support datasets using profile cards with demographic attributes, enabling it to learn seeker behaviors associated with each profile \cite{zhao-etal-2024-esc}. 

Despite conditioning on different seeker prompts, existing simulators inherit the base language model’s verbose and cooperative interaction bias. Moreover, even learned simulators tend to reflect only surface-level profile attributes, failing to induce deeper psychological traits across seeker types. Our simulator addresses these limitations.

\section{Data Construction and Feature Formulation}\label{sec:data-construction}

To build a population-diverse and controllable seeker simulator, we construct a large-scale dialogue dataset grounded in authentic online interactions as the training foundation.

\subsection{Data Collection and Preprocessing}\label{sec:3.1-data-collection}

We curated a dataset of real-world emotional support interactions from Reddit, specifically targeting subreddits with over 500,000 members, such as \texttt{r/offmychest} and \texttt{r/mentalhealth}. Reddit serves as an ideal source for training a feature-controllable simulator, as its in-the-wild nature encompasses a broad spectrum of informal and raw human expressions often absent in curated datasets.

Unlike traditional approaches that rely on heavily filtered datasets, we intentionally retained informal and raw elements. We included a wide variance in utterance lengths and aggressiveness, addressing the limitations of prior simulators that portray only articulate and cooperative seekers.

While prioritizing realism, we still implemented rigorous steps to ensure ethical integrity and basic data quality. Most importantly, all personally identifiable information (PII) was systematically masked to protect user anonymity.
We also discarded low-quality data based on conversation length, topics, and upvote counts.
The final dataset includes 11,066 dialogues. Detailed preprocessing procedures are provided in Appendix~\ref{app:preprocessing}.

\subsection{Feature Taxonomy for Seeker Profiles}\label{sec:feature-taxanomy}
To represent the multifaceted characteristics of seekers' conversational behavior, 
we define a comprehensive taxonomy consisting of nine distinct features. These are categorized into psychological and linguistic attributes. Detailed category definitions for each feature are provided in Appendix~\ref{app:feature-cat-definitions}.

\subsubsection{Psychological Features}
We incorporate six psychological features to model seeker behavior. 
The \textbf{main coping strategy} categorizes seekers into four types, grounded in the multidimensional coping framework~\cite{endler1994assessment}: \textit{problem\_focused}, \textit{emotion\_focused}, \textit{avoidant}, and \textit{maladaptive\_behavior}. The \textbf{utterance style} captures stylistic variation in seeker expression using three categories (\textit{plain}, \textit{upset}, \textit{verbose}), adapted from the PATIENT-$\Psi$ framework~\cite{wang-etal-2024-patient}. 
To model interactional dynamics, we define a three-level ordinal \textbf{resistance level} based on psychological reactance theory~\cite{brehm1981psychological}, and a three-level \textbf{engagement level} grounded in prior work on client participation in therapeutic settings~\cite{holdsworth2014client}.
Finally, we include a four-level \textbf{self-disclosure level} derived from Social Penetration Theory~\cite{altman1973social}, as well as \textbf{seeker reaction proportions} following prior work on modeling client reactions in emotional support dialogues~\cite{li-etal-2023-understanding}.


\subsubsection{Linguistic Features}
We further define three linguistic attributes that capture diverse structural patterns. The \textbf{verbosity level} is defined as a five-level feature based on the distribution of seeker utterance lengths.
We additionally include a binary \textbf{user profanity flag}, detected using the \texttt{profanity\_check} library. Finally, the \textbf{total dialogue turns level} captures the overall length of the interaction.

\subsection{Feature Annotation and Validation}\label{sec:3.3-feature-annotation}

To apply the feature taxonomy to our corpus, we employed a hybrid annotation framework combining LLM-based tagging and rule-based extraction.
\subsubsection{Psychological Feature Annotation}
Psychological features are annotated using an LLM-based tagging process, followed by targeted human validation on a subset of samples.

\paragraph{LLM-based Tagging Process}
We performed tagging at two different levels (dialogue and utterance level) and subsequently aggregated the annotations into dialogue-level representations. Detailed assignments are provided in Appendix~\ref{app:llm-based-psy-feature-tag}.


\paragraph{Human Validation} To verify the reliability of the LLM-based feature tagging, we conducted a validation study on a sample of 60 dialogues. Inter-annotator agreement (percent agreement) averages 0.57 across psychological features, while human–LLM alignment achieves 0.84 accuracy, indicating reliable alignment. Detailed validation procedures and feature-level validation results are provided in Appendix~\ref{app:llm-tagger-human-eval}.


\subsubsection{Linguistic Feature Extraction}
Linguistic features describe the measurable characteristics of the dialogue and are extracted using rule-based methods: \textit{verbosity level} from seeker utterance length, \textit{profanity flag} via \texttt{profanity-check}, and \textit{total dialogue turns level} from turn count. Detailed extraction rules are in Appendix~\ref{app:rule-based-ling-feature-tag}.

\subsection{Seeker Profile Construction}\label{sec:profile-construct}

We construct a final seeker profile by combining all annotated features with the seeker’s main problem. Formally, a seeker profile is represented as:
\[
P = [f_1, f_2, \dots, f_9, m],
\]
where each $f_i$ denotes an annotated \textbf{seeker feature} and $m$ represents the \textbf{seeker’s main problem}, both expressed in natural language. An example profile is in Figure~\ref{fig:seeker_profile_example}. The resulting dataset consists of 11,066 unique seeker profiles, split into 8,868 training, 1,094 validation, and 1,104 test profiles.
The seeker’s main problem is a summary of the first seeker utterance (i.e., original post) that describes the seeker's situation. The summaries are generated by GPT-4o-mini \cite{openai2024gpt4}, with detailed prompts provided in Table~\ref{tab:seeker-summary-prompt}.

\begin{figure*}[t]
  \centering
  \hspace*{-5mm}
  \includegraphics[width=1.05\textwidth]{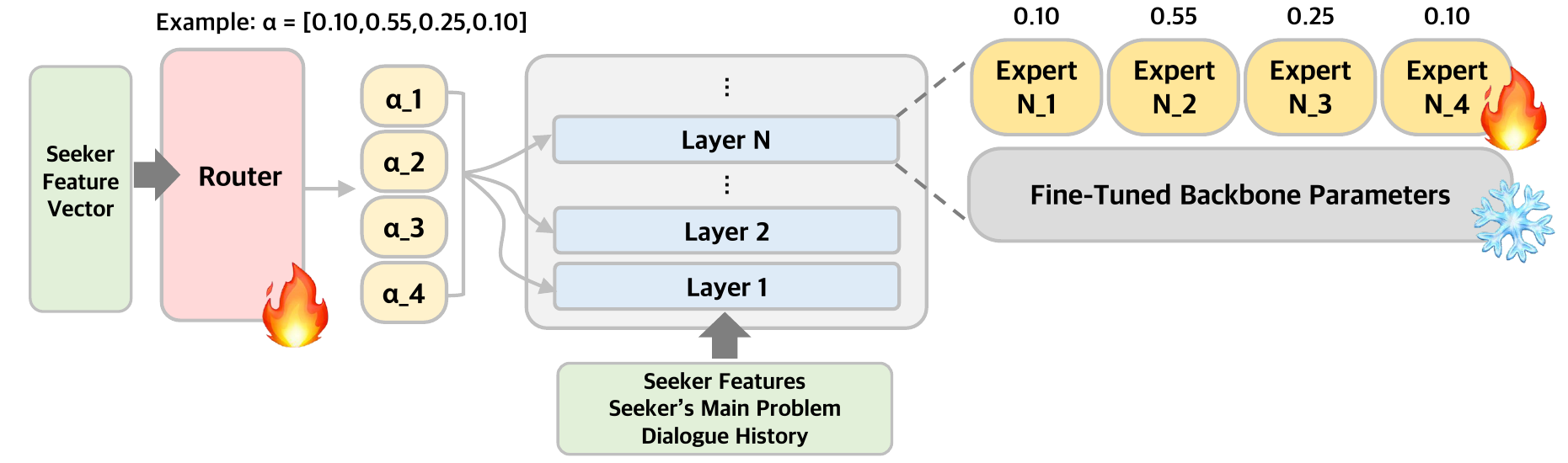}
  \caption{Overview of the MoE architecture. A frozen SFT backbone is augmented with multiple low-rank expert adapters at each linear layer. A shared routing network produces a dialogue-level routing vector $\boldsymbol{\alpha}$, which controls expert activation consistently across all layers.}
  \label{fig:2-moe-detail}
\end{figure*}

\section{Seeker Simulator Training}

To simulate diverse seeker behaviors in emotional support conversations, we employ a two-stage training framework: (1) \textbf{Supervised Fine-Tuning (SFT)} to establish a general conversational backbone, and (2) \textbf{Mixture-of-Experts (MoE) training} to enable controllable, attribute-specific generation. 

\subsection{Supervised Fine-Tuning}
\label{sec:backbone-sft}

The first stage establishes a linguistic foundation by capturing general conversational patterns of seekers. We use Llama-3-8B-Instruct as the base model and apply LoRA ($r=16$) to all linear layers. The model is trained to predict the next seeker utterance, given a seeker profile and dialogue history as a next-token prediction task. Upon convergence, the LoRA adapters are merged into the base model weights. This merged model is then entirely frozen, serving as a stable backbone for the subsequent expert-based training stage.

\subsection{MoE Training with Behavioral Routing}
\label{sec:moe-training}

SFT backbone often fails to manifest seeker behaviors faithful to the input seeker features since the feature signals are frequently overshadowed by long dialogue histories or system prompts.

To address this challenge, we introduce an explicit routing mechanism that decouples feature control from language modeling. Rather than expecting the model to infer seeker features from text, we provide a structured feature vector to control expert selection through a soft MoE formulation. An overview of the proposed framework is in Figure~\ref{fig:2-moe-detail}.

\subsubsection{Seeker Feature Vector Construction}

The routing network operates on a rule-based seeker feature vector $\mathbf{f} \in \mathbb{R}^{14}$ that encodes features associated with each seeker profile. This feature vector is constructed independently of the language modeling input and is used exclusively for routing, ensuring that expert selection is guided by structured behavioral signals. 

Categorical attributes (e.g., \textit{main coping strategy}) are mapped to one-hot representations for independent representation. Level-based attributes (e.g., \textit{resistance level}, \textit{engagement level}) are encoded as zero-centered normalized scalars to preserve their ordinal structure while maintaining comparable magnitudes across features.

\subsubsection{Shared Dialogue-Level Routing Network}

Given the seeker feature vector $\mathbf{f}$, we compute a routing distribution that determines how strongly each expert is activated. 

The routing network consists of $L$ residual MLP blocks followed by a gating layer.
Each block applies a standard residual update,
$\mathbf{h}_{\ell+1}
= \mathrm{LayerNorm}(\mathbf{h}_{\ell} + \mathrm{MLP}(\mathbf{h}_{\ell}))$,
and the final hidden vector $\mathbf{h}_L$ is mapped to

\begin{equation}
    \boldsymbol{\alpha} = \mathrm{softmax}(W_g \mathbf{h}_L) 
    \nonumber
\end{equation}
where $\boldsymbol{\alpha} \in \mathbb{R}^{N_E}$ and $N_E$ is the number of experts (we set $N_E=4$ and $L=3$). As shown in Figure~\ref{fig:2-moe-detail}, $\boldsymbol{\alpha}$ is computed once per dialogue session and shared across all layers. This global routing strategy ensures that behavioral differences between seekers are reflected consistently in expert activation throughout the entire generation process.

\subsubsection{LoRA-Based Experts}

Using the frozen SFT backbone from Section~\ref{sec:backbone-sft}, we attach $N_E$ low-rank expert adapters to each linear layer in both the attention and feed-forward (FFN) blocks as illustrated in Figure~\ref{fig:2-moe-detail}.
For an expert-augmented linear layer $\ell$, the output is computed as:
\begin{equation}
\mathbf{y}^{(\ell)} =
W^{(\ell)} \mathbf{x}^{(\ell)}
+
\sum_{i=1}^{N_E}
\alpha_i \cdot \Delta_i^{(\ell)}(\mathbf{x}^{(\ell)}),
\nonumber
\end{equation}
where $W^{(\ell)} \in \mathbb{R}^{d_{\mathrm{out}}^{(\ell)} \times d_{\mathrm{in}}^{(\ell)}}$ denotes the frozen backbone weight of layer $\ell$.
$\Delta_i^{(\ell)}(\mathbf{x}^{(\ell)})$ is the $i$th expert in layer $l$, parameterized as a LoRA adapter, i.e., $\mathbf{x}^{(\ell)} A_i^{(\ell)} B_i^{(\ell)}$, where $A_i^{(\ell)} \in \mathbb{R}^{d_{\mathrm{in}}^{(\ell)} \times r}$ and
$B_i^{(\ell)} \in \mathbb{R}^{r \times d_{\mathrm{out}}^{(\ell)}}$, with $r \ll d_{\mathrm{in}}^{(\ell)}, d_{\mathrm{out}}^{(\ell)}$ (we set $r=4$).


\subsubsection{Training Objectives}


The MoE model is trained using a combination of the standard language modeling loss and a Task-wise Decorrelation (TwD) loss, following the prior work of \citet{heterogeneous-mola}.

\paragraph{Task-wise Decorrelation Loss.}
To ensure that routing distributions reflect meaningful behavioral distinctions, we encourage samples with different feature labels to produce distinct routing vectors. 
For each sample $i$, the routing distribution $\boldsymbol{\alpha}^{(i)} \in \mathbb{R}^{E}$ is projected into a latent space via a learnable projection function:
\begin{equation}
\boldsymbol{\omega}^{(i)} = \phi(\boldsymbol{\alpha}^{(i)}),
\nonumber
\end{equation}
where $\phi(\cdot)$ denotes a learnable shared two-layer MLP that maps
$\boldsymbol{\alpha} \in \mathbb{R}^{N_E}$ to a latent embedding
$\boldsymbol{\omega} \in \mathbb{R}^{D}$, with $D=64$ and a GELU nonlinearity.

We then optimize the routing representations using a contrastive objective defined for each feature $f$:
\begin{equation}
\mathcal{L}_{\mathrm{TwD}}^{(f)} =
\mathbb{E}_{i}
\left[
-\log
\frac{\sum_{j \in \mathcal{P}_f(i)} \exp(s_{ij})}
{\sum_{k \neq i} \exp(s_{ik})}
\right],
\nonumber
\end{equation}
where $\mathcal{P}_f(i) = \{\, j \neq i \mid y_f^{(j)} = y_f^{(i)} \,\}$ denotes the set of positive samples sharing the same label for feature $f$ as sample $i$, $y_f^{(i)}$ is the discrete label of sample $i$ for feature $f$, and $s_{ij} = \boldsymbol{\omega}^{(i)} \cdot \boldsymbol{\omega}^{(j)} / \tau$ denotes cosine similarity with temperature $\tau$. The final TwD loss is obtained by averaging $\mathcal{L}_{\mathrm{TwD}}^{(f)}$ across all considered features.

\paragraph{Language Modeling Loss.}
We use a standard next-token prediction loss $\mathcal{L}_{\mathrm{LM}}$, which is computed exclusively on the next seeker utterance from seeker profile and dialogue history, following the same setup as in Section~\ref{sec:backbone-sft}.

\paragraph{Overall Objective.}
The final training objective is defined as:
\begin{equation}
\mathcal{L} = \mathcal{L}_{\mathrm{LM}} + \lambda \mathcal{L}_{\mathrm{TwD}}.
\nonumber
\end{equation}

Notably, we do not pre-assign semantic roles to individual experts. Instead, expert specialization emerges implicitly through joint optimization of the routing network and expert parameters. 
We analyze the resulting expert behaviors and routing patterns in Section~\ref{sec:mola-analysis}.

\section{Seeker Simulator Validation}\label{sec:seeker-validation}
We evaluate our seeker simulator across three dimensions for diverse and controllable simulation. Specifically, we evaluate \textbf{(1) profile adherence}: whether the generated utterances faithfully follow the intended seeker profiles; \textbf{(2) fidelity}: how well the simulator produces human-like and psychologically plausible utterances; and \textbf{(3) diversity}: whether the simulator can cover a broad, well-separated range of seeker populations. To isolate the performance of the seeker simulator, we fix the supporter model as GPT-5-mini, which demonstrated the highest overall performance in our supporter evaluation (Section \ref{sec:evaluate-supporter}).


\subsection{Profile Adherence}
To evaluate how accurately generated utterances reflect the nine-dimensional features specified in the input profile, we measure profile adherence.

\subsubsection{Experimental Setup}
\paragraph{Validation Pipeline and Metrics}
For each seeker profile, we generate a complete dialogue session with the GPT-5-mini supporter model. We then extract features from the generated seeker utterances using the LLM-based tagger and rule-based methods described in Section \ref{sec:3.3-feature-annotation} and compare them with the input features using the macro F1-score.


\paragraph{Baselines}\label{sec:profile_adherence_baselinesimulators}
We use a range of baselines that represent different levels of model capability and training methodologies.
\begin{itemize}
\item Zero-shot Base Models: GPT-4.1-mini, Llama-3-8B-Instruct~\cite{touvron2024llama3}, Qwen-2.5-14B-Instruct~\cite{qwen2024technical}, GPT-5~\cite{singh2025openaigpt5card}, and DeepSeek-V3.2~\cite{deepseekai2025deepseekv32pushingfrontieropen} 
\item SFT: We fine-tune Llama-3-8B-Instruct on our curated Reddit dialogue dataset using standard next-token prediction. Specifically, given a seeker profile and dialogue history, the model is trained to predict the next seeker utterance.
\item Contrastive Learning: A variant of the SFT model that incorporates disentanglement loss to better distinguish between different feature levels. Detailed training objectives and implementation details are provided in Appendix~\ref{app:cl}.

\end{itemize}

\subsubsection{Results and Analysis}
\label{sec:mola-analysis}
The experimental results, summarized in Table~\ref{tab:profile_adherence_f1} and~\ref{tab:all_feature_macro_f1}, reveal several key insights regarding the controllability of seeker simulators. Linguistic features such as \textbf{verbosity} and \textbf{dialogue length} are relatively easy to learn, with all training-based methods substantially outperforming zero-shot baselines. In contrast, psychological features like \textbf{resistance} and \textbf{self-disclosure} are much harder: zero-shot models score around 0.2 in Macro F1, standard training reaches only around 0.3, and only our MoE-based model exceeds 0.4. Overall, our model achieves the highest profile adherence across all baselines, including standard zero-shot approaches, frontier reasoning models, and other training-based methods. Feature-level accuracy and correlation scores are in Table~\ref{tab:all_feature_macro_f1} and~\ref{tab:feature_profile_correlations}.


\begin{table}[t]
\centering
\small
\setlength{\tabcolsep}{4pt}
\renewcommand{\arraystretch}{1.1}
\begin{tabular}{l|cccc}
\hline
\textbf{Simulator} 
& \textbf{Mean} $\uparrow$ 
& \textbf{Std} $\downarrow$ 
& \textbf{Min} $\uparrow$ 
& \textbf{Max} $\uparrow$ \\
\hline
GPT-4.1-mini          & 0.301 & 0.131 & 0.160 & 0.580 \\
Llama-3-8B-Instruct   & 0.259 & 0.148 & 0.110 & 0.580 \\
Qwen-2.5-14B-Instruct & 0.284 & \textbf{0.095} & 0.150 & 0.470 \\
GPT-5                 & 0.319 & 0.216 & 0.150 & 0.840 \\
DeepSeek-V3.2         & 0.431 & 0.218 & 0.180 & 0.910 \\
SFT                   & 0.515 & 0.160 & 0.360 & 0.760 \\
Contrastive Learning  & 0.484 & 0.178 & 0.340 & \textbf{0.850} \\
\textbf{Ours}         & \textbf{0.549} & 0.125 & \textbf{0.430} & 0.740 \\
\hline
\end{tabular}
\caption{Profile adherence results measured by Macro F1. 
$\uparrow$ indicates that higher values are better, while $\downarrow$ indicates that lower values are better.}
\label{tab:profile_adherence_f1}
\end{table}


\paragraph{Analysis on MoE's Adaptive Routing}

To identify the mechanism behind improved profile adherence, we analyze whether individual experts specialize in distinct behavioral roles. 

Specifically, we interpret expert behavior by first assigning each dialogue to a \textit{dominant expert}, defined as $e^* = \arg\max \boldsymbol{\alpha}$, where $\boldsymbol{\alpha}$ is the routing vector. We visualize the routing distributions $\boldsymbol{\alpha}$ using PCA (Figure~\ref{fig:appendix-alpha-pca}), where each point is colored by its dominant expert. The clear separation of clusters indicates that the routing network partitions the representation space into distinct regions, providing a structural basis for expert-level interpretation. 

We then analyze the conditional distributions $P(\text{feature} \mid e^*)$ over these assignments. These distributions are visualized as row-normalized heatmaps in Figure~\ref{fig:appendix-expert-heatmaps}.  As an example, Expert 1 shows strong concentration on high engagement and self-disclosure, along with lower resistance in the heatmap (e.g., engagement: 1.0, resistance: 0.76, self-disclosure level 4: 0.53), corresponding to a highly cooperative and open behavioral pattern.

The post-training routing analysis demonstrates that our feature-aware MoE mechanism successfully induces behaviorally meaningful specialization, aligned with distinct seeker profiles:

\begin{itemize}
    \item \textit{Expert 0 (Emotion-Oriented):} Specializes in seekers with high emotional distress, characterized by emotion-processing coping strategy, high resistance, and expressive (\textit{upset}/\textit{verbose}) communication.
    
    \item \textit{Expert 1 (Collaborative \& Open):} Captures highly cooperative dynamics, focusing on seekers with high engagement, low resistance, and high self-disclosure.
    
    \item \textit{Expert 2 (Pragmatic \& General):} Exhibits relatively weaker specialization, primarily handling problem-focused seekers with "average" feature levels.
    
    \item \textit{Expert 3 (Reclusive):} Manages seekers who maintain psychological distance, utilizing avoidant or maladaptive strategies with low engagement and minimal disclosure.
    
\end{itemize}

This suggests that our proposed MoE framework induces the emergent specialization of intrinsic parameter subspaces, effectively partitioning behavioral counseling regimes. Detailed figures are provided in Appendix~\ref{app:moe-analysis}.

\subsection{Fidelity}
Fidelity focuses on whether the generated utterances reflect the authentic communicative patterns of human help-seekers. To assess this, we conduct an expert evaluation comparing our simulator with several representative baselines. All evaluations were conducted by three graduate students in clinical psychology who were trained on the evaluation instructions for each criterion.

\subsubsection{Experimental Setup}
\paragraph{Evaluation Task} For each trial, experts were presented with two dialogues generated by different seeker simulators starting from the same initial problem summary. Experts select the dialogue whose seeker utterances better satisfy each evaluation criterion, or choose "Tie" if the quality is indistinguishable. Experts evaluate seeker fidelity along three dimensions: \textbf{(1) Linguistic Naturalness}, \textbf{(2) Role Authenticity}, and \textbf{(3) Psychological Plausibility}, which respectively assess language fluency, role consistency, and emotional coherence. Detailed definitions, instructions, annotator information, and inter-annotator agreement are provided in Appendix~\ref{app:expert-eval-fidelity} and Table~\ref{tab:expert-eval-instruction}.

\paragraph{Data Sampling and Session Control} We randomly sample 90 unseen seeker summaries from Reddit and fix all dialogue sessions to 10 turns to control for length effects.

\paragraph{Baseline Simulators}\label{sec:fidelity_baselinesimulators}
We compare our model against three representative seeker simulators. To ensure that each baseline operates at its intended capacity, we adopt the original profile structures and prompting schemes specified in their respective papers. Specifically, we incorporate the seeker simulator in ESC-Judge \cite{madani-srihari-2025-esc}, a zero-shot baseline utilizing GPT-4o without additional training, and ESC-Role \cite{zhao-etal-2024-esc} and Eeyore \cite{liu-etal-2025-eeyore}, both of which are fine-tuned on emotional support datasets.

\subsubsection{Results and Analysis}
The results of the expert evaluation are summarized in Table~\ref{tab:fidelity_winrate}. Overall, our simulator consistently outperformed all baseline models across all three dimensions, achieving an average win rate of 69.5\% across all pairwise comparisons and criteria.


\begin{table}[t]
\centering
\normalsize
\setlength{\tabcolsep}{5pt}
\renewcommand{\arraystretch}{1.1}
\resizebox{\columnwidth}{!}{%
\begin{tabular}{l|ccc}
\toprule
\textbf{Comparison} & \thead{\textbf{Linguistic}\\\textbf{Naturalness}} & \thead{\textbf{Role}\\\textbf{Authenticity}} & \thead{\textbf{Psychological}\\\textbf{Plausibility}} \\
\midrule
Ours vs. Eeyore    & 68.9 & 66.7 & 71.1 \\
Ours vs. ESC-Judge & 68.9 & 72.2 & 62.2 \\
Ours vs. ESC-Role  & 80.0 & 67.8 & 67.8 \\
\bottomrule
\end{tabular}%
}
\caption{Win rates (\%) of our simulator against baseline simulators across three fidelity criteria. Win rate is computed as the proportion of wins out of total comparisons, including ties.}
\label{tab:fidelity_winrate}
\end{table}

Specifically, our model showed a distinct advantage in linguistic naturalness over Eeyore and ESC-Role, capturing more raw communicative patterns found in real-world emotional support interactions. Furthermore, compared to ESC-Judge, our simulator demonstrated higher role authenticity, as individuals often exhibit uncertainty about their own problems rather than producing overly structured or verbose explanations. Most significantly, the superior performance in psychological plausibility validates that our simulator captures authentic patterns of emotional change, maintaining a coherent emotional trajectory across the dialogue. Detailed win/loss/tie breakdowns are provided in Table~\ref{tab:human-pref-metric}.

\subsection{Diversity}

\subsubsection{Experimental Setup} To assess each simulator's coverage of diverse seeker characteristics, we evaluate the diversity of the generated utterances. Using 300 held-out profiles per simulator, we conduct dialogues with GPT-5-mini as the supporter. We compare our simulator against four baselines: ClientCAST \cite{wang2024clientcenteredassessmentllmtherapists}, ESC-Judge, ESC-Role, and Eeyore. Details on simulator-specific profile construction are in Appendix~\ref{app:profile-construction-each}. For each dialogue, seeker utterances are aggregated into a single embedding using the all-MiniLM-L6-v2 to capture the seeker’s overall expressive pattern \cite{reimers-gurevych-2019-sentencebert}.

\subsubsection{Evaluation Metrics} We assess diversity using a combination of visualization-based analysis and lexical, semantic, and sentiment metrics. Detailed metric definitions and formulations are provided in Appendix~\ref{app:diversity-metrics}.

\subsubsection{Results and Analysis}
\begin{figure}[t]
  \centering
  \includegraphics[width=\columnwidth]{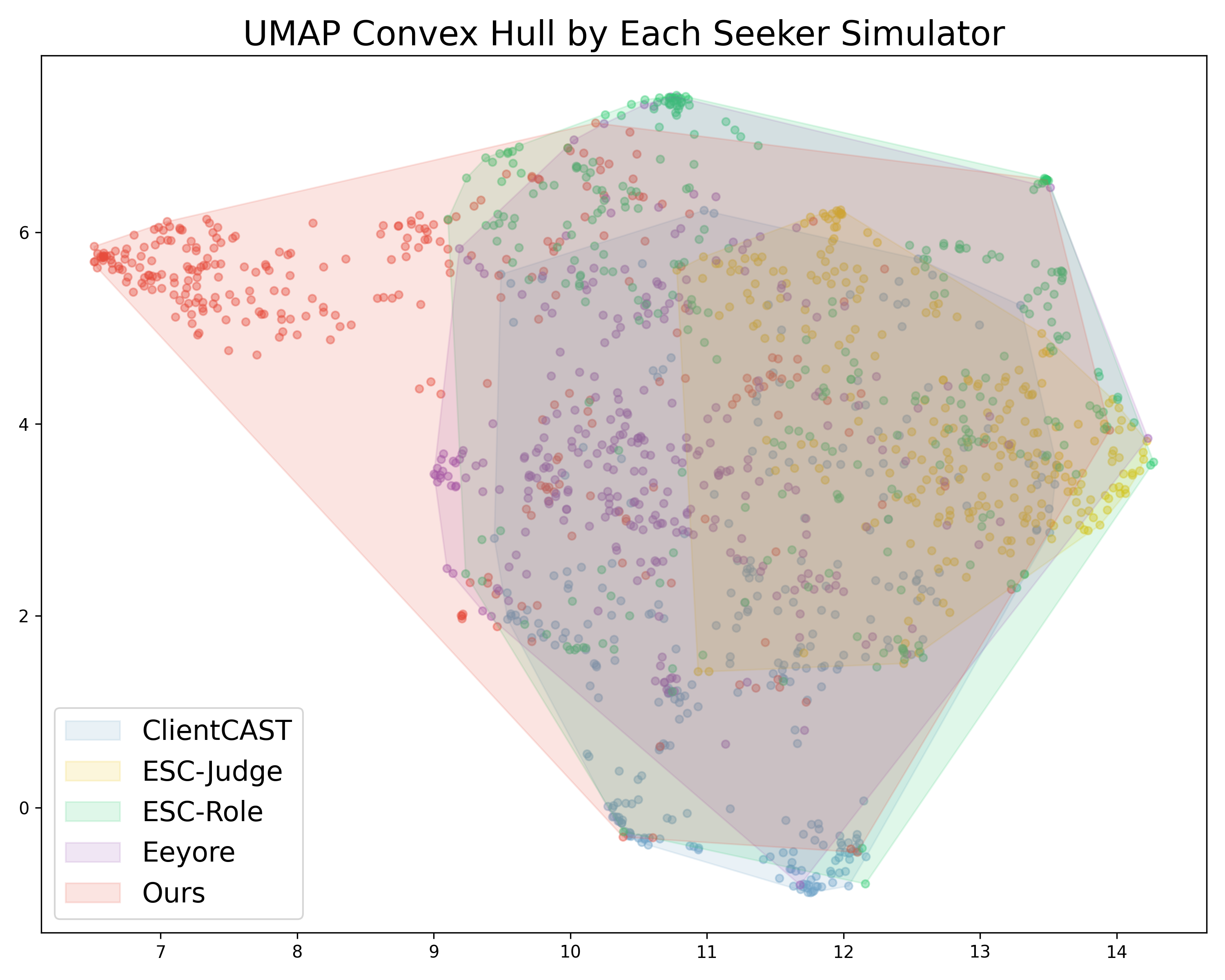}
  \caption{UMAP projection of dialogue-level seeker embeddings across simulators.}
  \label{fig:diversity-umap}
\end{figure}

Figure~\ref{fig:diversity-umap} visualizes the semantic diversity of seeker utterances using UMAP projections. Our simulator exhibits substantially broader coverage compared to all baseline simulators, indicating a wider range of expressed behaviors and interaction styles. In contrast, baseline simulators tend to concentrate within narrower regions of the embedding space, suggesting limited behavioral diversity. Notably, simulators without fine-tuning (ESC-Judge and ClientCAST) occupy the most constrained regions, while SFT-based simulators (ESC-Role and Eeyore) achieve moderately broader coverage. Our MoE-based simulator achieves the largest coverage area, demonstrating that assigning dedicated experts to distinct seeker profiles effectively expands behavioral diversity.

\subsubsection{Additional Quantitative Analysis}
Beyond visualization, we further assess diversity using lexical, semantic, and sentiment-based metrics. Across all metrics, our simulator consistently demonstrates higher diversity than baseline models, which exhibit more repetitive language and narrower semantic and affective ranges. Detailed quantitative results are provided in Table~\ref{tab:seeker_diversity_extended}.



\section{Supporter Model Evaluation}\label{sec:evaluate-supporter}

In this section, we utilize our controllable seeker simulator to evaluate the performance of state-of-the-art supporter models, examining how their performance and supportive behaviors vary across diverse seeker populations. To examine the impact of seeker populations on evaluation outcomes, we evaluate the same support models using other existing seeker simulators under identical settings.

\subsection{Evaluation Setup}\label{sec:6.1eval-setup}
We adopt a unified evaluation pipeline across all seeker–supporter pairs, applying the same dialogue generation and evaluation rules. All evaluations are based on multi-turn dialogues between the seeker simulator and the supporter model, followed by automated evaluation across ten distinct metrics.


Each dialogue begins with an initial seeker utterance based on the seeker’s main problem. For each seeker simulator, we use 300 held-out test profiles. Dialogues terminate either via closing exchanges or, in our simulator, through explicit termination modeling using a learned \texttt{<|end\_of\_dialogue|>} token conditioned on the total dialogue turns level. The maximum dialogue length is set to 20 turns. A consistent supporter role prompt is used for all supporter models. Detailed evaluation settings and prompt templates are provided in Appendix~\ref{app:evaluation-details-for-supporter-eval}.

\subsection{Evaluation Metrics}

To measure the supportive capabilities of each supporter model, we adopt the expert-validated framework from \citet{kim-etal-2025-dialogue}. Following the original study, all metrics are evaluated using GPT-4o-mini and are categorized into the following three groups for clarity: (1) ES Skills (Identification, Comforting, Suggestions, Experience, Informativeness), (2) General Skills (Consistency, Role-Adherence, Expression, Humanness), and (3) Overall.

\subsection{Supporter Models}
We evaluate a diverse range of supporter models to identify generalizable performance trends.

\begin{itemize}
    \item Prompt-based: GPT-5-mini, Llama-3-8B-Instruct.
    \item Fine-tuned: Llama-3-8B-Instruct models fine-tuned on emotional support datasets, i.e., Llama-ESConv, Llama-ExTES, Llama-Psych8k, Llama-PsyInsight, and Llama-CounselChat (see Appendix~\ref{app:supporter-train-datasets} for dataset sources).
\end{itemize}

\subsection{Evaluation Results}
Using the unified evaluation setup in Section~\ref{sec:6.1eval-setup}, we analyze how supporter performance varies across different seeker populations, highlighting the importance of population-diverse evaluation.

First, supporter model rankings vary substantially across seeker simulators. With cooperative simulators such as ESC-Judge and ESC-Role, models achieve uniformly high ES skills with small gaps, whereas under our simulator's more challenging seeker behaviors, ES scores drop sharply and model rankings change noticeably. This suggests that evaluations based on limited seeker populations can overestimate supporter robustness.

Second, evaluation metrics differ in their sensitivity to seeker behaviors. While general fluency remains relatively stable across simulators, ES skills—particularly suggestions and informativeness—exhibit larger performance drops and higher variance under our simulator. This indicates that challenging seeker behaviors primarily affect a model’s ability to deliver substantive support rather than surface-level language quality (Table~\ref{tab:merged_metrics_10cols}). 


Third, qualitative analysis of low-scoring dialogues reveals two recurring failure patterns. With resistant seekers, models fall back on repetitive apologies instead of providing substantive empathy or constructive suggestions. With low-engagement seekers, they persist in monotonous probing rather than adapting strategies like validation or reflection. These patterns explain the observed drops in Suggestions and Informativeness.

\subsection{Validity of Automatic Evaluation}
\label{sec:automatic_evaluation_validity}

To verify that the observed score drops across seeker simulators reflect genuine deficiencies in the supporter model's skills, rather than the automated judge assigning low scores in response to the seeker's negative or resistant behaviors, we conduct a human evaluation. Two graduate students in clinical psychology evaluated 60 dialogues using the same 1--5 rating scale and rubrics as the automated evaluation (details in Appendix~\ref{app:human_eval_details}).

Specifically, we compute $\Delta_\text{Human}$ and $\Delta_\text{LLM}$ as the mean score differences (\textit{Ours} $-$ \textit{ESC-Judge}) under human and automated evaluation, respectively. A negative $\Delta$ indicates that supporter models perform worse when evaluated with our simulator compared to ESC-Judge. As shown in Table~\ref{tab:human_eval}, both $\Delta_\text{Human}$ and $\Delta_\text{LLM}$ are consistently negative across all five ES-skills, confirming that the score drops are not an artifact of automated scoring. Notably, with the exception of Experience Sharing, $\Delta_\text{Human}$ is larger in magnitude than $\Delta_\text{LLM}$ across all metrics, indicating that human experts perceive the performance gap even more strongly than the automated judge. This suggests that the score drops observed under our simulator reflect genuine supporter performance degradation, rather than systematic bias introduced by the seeker's negative behaviors.

\begin{table}[t]
\centering
\small
\setlength{\tabcolsep}{4pt}
\renewcommand{\arraystretch}{1.1}
\begin{tabular}{l|ccc}
\toprule
\textbf{Metric} & $\boldsymbol{\Delta_\text{Human}}$ & $\boldsymbol{\Delta_\text{LLM}}$ & \textbf{Spearman $\rho$} \\
\midrule
Identification   & -1.167 & -0.733 & 0.454 \\
Comforting       & -0.900 & -0.833 & 0.377 \\
Suggestions      & -1.583 & -1.067 & 0.744 \\
Exp. Sharing     & -0.417 & -1.067 & 0.402 \\
Informativeness  & -1.917 & -1.467 & 0.673 \\
\bottomrule
\end{tabular}
\caption{Score differences ($\Delta$ = Ours $-$ ESC-Judge) under human and automated evaluation, respectively, and their rank correlations (Spearman $\rho$) across ES-skills. All correlations are significant ($p < 0.05$).}
\label{tab:human_eval}
\end{table}

\subsection{Guidelines for Seeker Profile Configuration}


Developers of emotional support chatbots can configure input seeker profiles according to the target seeker population of the chatbot. To illustrate how our framework can be applied in practice, we apply the same feature annotation pipeline to a real-world spoken counseling dialogue dataset \cite{cpts2} and compare feature distributions between this target seeker population and our Reddit-based training data (Figure~\ref{fig:realworld-feature-compare-all}). These population-level differences offer a concrete basis for configuring target seeker profiles. For instance, a chatbot targeting general online users may configure profiles based on Reddit-based interactions, where seekers tend to communicate in a plain and concise style with varying levels of self-disclosure. In contrast, a chatbot targeting formal clinical psychotherapy settings may prioritize verbose utterance styles and lower self-disclosure levels (levels 1--2), reflecting the more structured and surface-level communication patterns observed in spoken counseling interactions.

\section{Conclusion}
\label{sec:conclusion}
We propose an evaluation framework for emotional support models based on a controllable seeker simulator, enabling assessment under diverse emotional support contexts. Our simulator demonstrates strong profile adherence, fidelity, and diversity, providing a reliable foundation for evaluation. Using this framework, we show that supporter performance varies significantly across seeker populations, revealing that conventional evaluations can mask meaningful weaknesses and overestimate real-world support effectiveness.

\section*{Limitations}
\label{sec:limitations}

Our work demonstrates that a controllable and population-diverse seeker simulator enables systematic evaluation of emotional support models across a wide range of interaction settings. However, our framework evaluates emotional support within a single dialogue, and therefore does not capture the cumulative nature of emotional support effects, which often unfold across repeated interactions in real-world settings. 

In addition, we represent seekers using profiles defined by combinations of behavioral features to enable controlled comparison. While this abstraction robustly captures diverse seeker characteristics, some features may naturally evolve over the course of an interaction. Although such variation may be partially reflected in the training data, dynamically changing features are treated as fixed during evaluation, and finer-grained control over temporal feature transitions remains an open direction.

Finally, our evaluation prioritizes supporting skills and overall conversational quality, using ES skills and general skills to assess supporter performance under diverse seeker populations. While this reveals meaningful performance differences across populations, it does not directly measure longer-term outcomes such as sustained emotional recovery, well-being, or behavioral change. Incorporating complementary evaluation settings and outcome-oriented metrics remains a promising direction for future extensions of this framework.

\section*{Acknowledgments}
\label{sec:acknowledgments}
This work was supported by the Creative-Pioneering Researchers Program through Seoul National University and by the National Research Foundation of Korea (NRF) grants (RS-2024-00333484 and RS-2024-00414981) funded by the Korean government (MSIT).

We used AI assistants to proofread the writing and to help with coding.

\bibliography{custom}


\newpage
\appendix


\appendix
\section{Data Preprocessing Details}\label{app:preprocessing}
In addition to the general data quality criteria described in the main text, we applied several concrete preprocessing procedures to the dataset.

We retained only dialogues consisting of at least five turns to ensure sufficient interactional context for modeling seeker behavior. Topic-based filtering was performed using GPT-4o-mini to exclude threads unrelated to mental health, such as requests for investment advice or restaurant recommendations. To ensure stable model training, dialogues containing non-English utterances were removed. We also removed emojis and other non-standard symbolic characters to reduce noise and variability in textual representations. Finally, a minimum upvote threshold was applied to filter out low-quality or low-engagement content.

These steps were applied uniformly across the dataset prior to training.

\section{Feature Category Definitions}
\label{app:feature-cat-definitions}
To enable fine-grained control over diverse seeker behaviors, we define a set of fine-grained categories for each of the nine features in our seeker profile taxonomy.
These categories are designed to capture subtle yet practically meaningful variations in how seekers express emotions, engage with counselors, and respond throughout emotional support dialogues.
The category-level definitions for each feature are provided below.

\subsection{Coping Strategy}
\begin{itemize}
  \item \textbf{problem\_focused}: Engages in concrete actions or cognitive efforts (e.g., planning, evaluating options, or using external resources) to address or change the stressor.
  \item \textbf{emotion\_processing}: Focuses on expressing, exploring, or emotionally processing feelings rather than directly solving the problem.
  \item \textbf{avoidant}: Deliberately avoids thinking about, confronting, or engaging with the problem or its associated emotions.
  \item \textbf{maladaptive\_behavior}: Responds to stress through currently ongoing self-destructive or harmful behaviors.
\end{itemize}

\subsection{Utterance Style}
While the original framework suggests a broader range of styles, we focus on the three most prevalent styles identified in our Reddit corpus—\textit{plain}, \textit{upset}, and \textit{verbose}—excluding those with negligible frequency in online support settings.
\begin{itemize}
  \item \textbf{plain}: Communicates thoughts and answers in a direct, concise, and emotionally neutral manner.
  \item \textbf{upset}: Expresses strong negative emotions such as anger or frustration in a confrontational or resistant tone.
  \item \textbf{verbose}: Produces excessively long and detailed utterances that disrupt conversational flow.
\end{itemize}

\subsection{Resistance Level}
\begin{itemize}
  \item \textbf{high}: Displays persistent and pervasive resistance across most topics and counselor interventions throughout the dialogue.
  \item \textbf{medium}: Shows situational resistance triggered by specific topics or interventions, alternating with periods of cooperation.
  \item \textbf{low}: Is generally cooperative, with resistance absent or limited to brief and minor instances.
\end{itemize}

\subsection{Engagement Level}
\begin{itemize}
  \item \textbf{high}: Actively participates as an equal partner by reflecting, expressing emotions, and driving the conversation forward.
  \item \textbf{medium}: Participates inconsistently, with engagement limited to specific dimensions or topics.
  \item \textbf{low}: Demonstrates minimal willingness to engage, providing short, dismissive, or disengaged responses.
\end{itemize}

\subsection{Self-Disclosure Level}
\begin{itemize}
  \item \textbf{1 (orientation)}: Shares only surface-level, socially normative information without emotional content or personal meaning.
  \item \textbf{2 (exploratory affective exchange)}: Shares personal facts or light emotions without deep vulnerability.
  \item \textbf{3 (affective exchange)}: Reveals emotionally meaningful experiences, fears, or vulnerabilities tied to personal significance.
  \item \textbf{4 (stable exchange)}: Expresses identity-level beliefs, core assumptions, or existential conclusions that generalize beyond specific events.
\end{itemize}

\subsection{Client Reaction Proportions}
\begin{itemize}
  \item \textbf{positive}: Responds to the counselor with openness, cooperation, or constructive engagement.
  \item \textbf{neutral}: Responds without clear acceptance or rejection, remaining vague or minimally responsive.
  \item \textbf{negative}: Pushes back against, rejects, or deflects the counselor's message in a resistant or defensive manner.
\end{itemize}

\subsection{Profanity Flag}
\begin{itemize}
  \item \textbf{true}: Contains profanity or explicit offensive language.
  \item \textbf{false}: Does not contain profanity or explicit offensive language.
\end{itemize}

\subsection{Verbosity Level}
\begin{itemize}
  \item \textbf{very\_short}: Fewer than 15 tokens.
  \item \textbf{short}: 15--29 tokens.
  \item \textbf{medium}: 30--59 tokens.
  \item \textbf{long}: 60--99 tokens.
  \item \textbf{very\_long}: 100 tokens or more.
\end{itemize}

\subsection{Total Dialogue Turns Level}
\begin{itemize}
  \item \textbf{short}: 4--5 turns.
  \item \textbf{medium}: 6--8 turns.
  \item \textbf{long}: 9 turns or more.
\end{itemize}

\section{LLM-based Psychological Feature Tagging}
\label{app:llm-based-psy-feature-tag}
To capture both stable interaction patterns and turn-level variations, we perform feature tagging at two levels—dialogue level and utterance level—and subsequently aggregate the annotations into dialogue-level representations used for model training and evaluation.

\begin{itemize}
\item \textbf{Dialogue-level Features:} \textit{Main coping strategy}, \textit{utterance style}, \textit{resistance level}, and \textit{engagement level} are annotated once per dialogue. Since these features reflect the seeker’s dominant interaction patterns, the LLM processes the entire dialogue history to infer a single representative label. See Table~\ref{tab:dialogue_level_raw_prompts} for detailed prompt specifications.

\item \textbf{Utterance-level Features:} In contrast, \textit{self-disclosure level} and \textit{seeker reaction proportions} may vary across turns depending on the conversational context including preceding supporter responses. Accordingly, these features are tagged at each turn based on supporter–seeker utterance pairs to capture such dynamic shifts. To construct final dialogue-level profile, \textit{self-disclosure level} scores are averaged and rounded, while \textit{seeker reaction proportions} is summarized as a distribution of positive, neutral, and negative responses. The common system instruction is provided in Figure~\ref{fig:common-feature-tag-instruction}, and detailed prompt specifications are listed in Table~\ref{tab:turn_level_raw_prompts}.
\end{itemize}

\begin{figure}[htbp]  
\centering  
\begin{tcolorbox}[
  colback=gray!5,
  colframe=gray!50,
  boxrule=0.6pt,
  arc=3pt,
  left=6pt,
  right=6pt,
  top=6pt,
  bottom=6pt,
  width=\columnwidth
]
\textbf{Common system instruction}

\smallskip
\raggedright
\small
\ttfamily
You are an expert evaluator. You will be provided with the client's main problem for context, followed by a multi-turn dialogue between a counselor and that client.
\end{tcolorbox}
\caption{Common system instruction for feature tagging.}
\label{fig:common-feature-tag-instruction}  
\end{figure}


\section{Human Validation for LLM-based Tagging}
\label{app:llm-tagger-human-eval}

To enhance annotation consistency, evaluators were randomly assigned to three specialized groups based on feature characteristics:

\begin{itemize}
\item \textbf{Group A}: Evaluates psychological and expressive characteristics (\textit{main coping strategy}, \textit{utterance style}).
\item \textbf{Group B}: Evaluates interactional attitudes (\textit{resistance level}, \textit{engagement level}).
\item \textbf{Group C}: Evaluates turn-level dynamics (\textit{self-disclosure level}, \textit{seeker reaction}).
\end{itemize}

Feature-level inter-annotator agreement (IAA) and human–LLM alignment results are reported in Table~\ref{tab:iaa_alignment}.

\begin{table}[t]
\centering
\small
\renewcommand{\arraystretch}{1.15}
\setlength{\tabcolsep}{6pt}
\begin{tabular}{l cc}
\hline
\textbf{Feature} & \textbf{IAA} & \textbf{Human--LLM} \\
                 & \textbf{(Percent Agr.)} & \textbf{(Accuracy)} \\
\hline
Main coping strategy   & 0.442 & 0.729 \\
Utterance style        & 0.528 & 0.852 \\
Resistance level       & 0.694 & 0.902 \\
Engagement level       & 0.597 & 0.854 \\
Self-disclosure level  & 0.564 & 0.827 \\
Seeker reaction        & --    & 0.860 \\
\hline
\end{tabular}
\caption{Feature-level inter-annotator agreement and human--LLM alignment.}
\label{tab:iaa_alignment}
\end{table}




\section{Rule-based Linguistic Feature Extraction}
\label{app:rule-based-ling-feature-tag}
\subsection{Linguistic Feature Extraction Rules}

Linguistic features are extracted using deterministic, rule-based procedures.
The \textit{verbosity level} is computed as the average token count of seeker utterances, excluding the initial turn corresponding to the original Reddit post, and discretized into a five-level scale based on the empirical distribution of the dataset.
The \textit{user profanity flag} is extracted using the \texttt{profanity-check} library and assigned as a binary value, where the flag is set to 1 if any seeker utterance exceeds a predefined probability threshold for profanity, and 0 otherwise.
The \textit{total dialogue turns level} is derived from the total number of turns in the dialogue and categorized into three tiers—Short, Medium, and Long—according to the dataset’s distribution.

These structural features provide objective signals that complement the psychological annotations by reflecting surface-level behavior and overall dialogue complexity.



\section{Implementation Details for Contrastive Learning}
\label{app:cl}
This section provides the mathematical formulation of the proposed \textbf{Disentanglement Loss ($L_D$)} and the technical specifications for pseudo-feature generation in contrastive learning baseline.

\subsection*{Training Objective}
The model is trained by combining the standard language modeling loss ($L_{LM}$) with our proposed Disentanglement Loss ($L_D$). The total training objective is defined as:
$$L_{total} = L_{LM} + \lambda_{D} L_D$$

$L_{LM}$ denotes the standard cross-entropy loss on the original dataset. $L_D$ is designed to ensure that the model generates responses that are clearly distinguishable based on the provided seeker features. The formulation of $L_D$ is as follows:

\begin{equation}
L_{D}
= -\log \frac{P(y \mid x_o)}
{P(y \mid x_o) + \sum_{i=1}^{N} P(y \mid x_{p,i})}
\nonumber
\end{equation}

where:
\begin{itemize}
    \item $x_{o}$: The original input.
    \item $x_{p,i}$: The $i$-th pseudo-input where features are intentionally modified.
    \item $y$: The ground-truth next client utterance from the original data.
    \item $N$: The number of pseudo-samples generated per anchor (set to $N=3$ in this study).
\end{itemize}

By maximizing the log-likelihood of the ground-truth response under the original features relative to the modified features, $L_D$ forces the model to be highly sensitive to the specific feature descriptions in the system prompt.

\subsection*{Pseudo-feature Generation}
To construct the contrastive set for $L_D$, we strategically manipulate the \texttt{seeker\_features} block within the system prompt. For each sample, three pseudo-variants are generated using the following priority rules:
\begin{itemize}
    \item \textbf{Extreme Flip}: Numerical or ordinal features such as \textit{engagement level}, \textit{resistance level}, and \textit{self-disclosure level} are flipped to their opposite extremes (e.g., \textit{high} $\leftrightarrow$ \textit{low}).
    \item \textbf{Categorical Shift}: Categorical features like \textit{main coping strategy} or \textit{utterance style} are replaced with a randomly selected alternative category.
    \item \textbf{Medium Flip}: Neutral values (e.g., \textit{medium}) are randomly reassigned to extreme values to enhance the discriminative signal.
\end{itemize}

\subsection*{Training Configuration}
The model was trained using the hyperparameters summarized in Table \ref{tab:hyperparams}.

\begin{table}[h]
\centering
\caption{Training Configuration for CL}
\label{tab:hyperparams}
\small 
\renewcommand{\arraystretch}{1.2} 
\begin{tabularx}{\columnwidth}{lX} 
\hline
\textbf{Category} & \textbf{Configuration} \\ \hline
Base Model & Meta-Llama-3-8B-Instruct \\
LoRA $r$ / $\alpha$ & 16 / 16 \\
LoRA Target & All Linear Layers \\
LoRA Dropout & 0.05 \\
Learning Rate & $2 \times 10^{-5}$ \\
Epochs & 1 \\
Batch Size & 16 (4 per device $\times$ 4 GA) \\
Max Length & 2048 \\
$\lambda_D$ & 1.0 \\ \hline
\end{tabularx}
\end{table}

\section{Routing Analysis in MoE}
\label{app:moe-analysis}

\begin{figure}[h]
    \centering
    \includegraphics[width=0.8\linewidth]{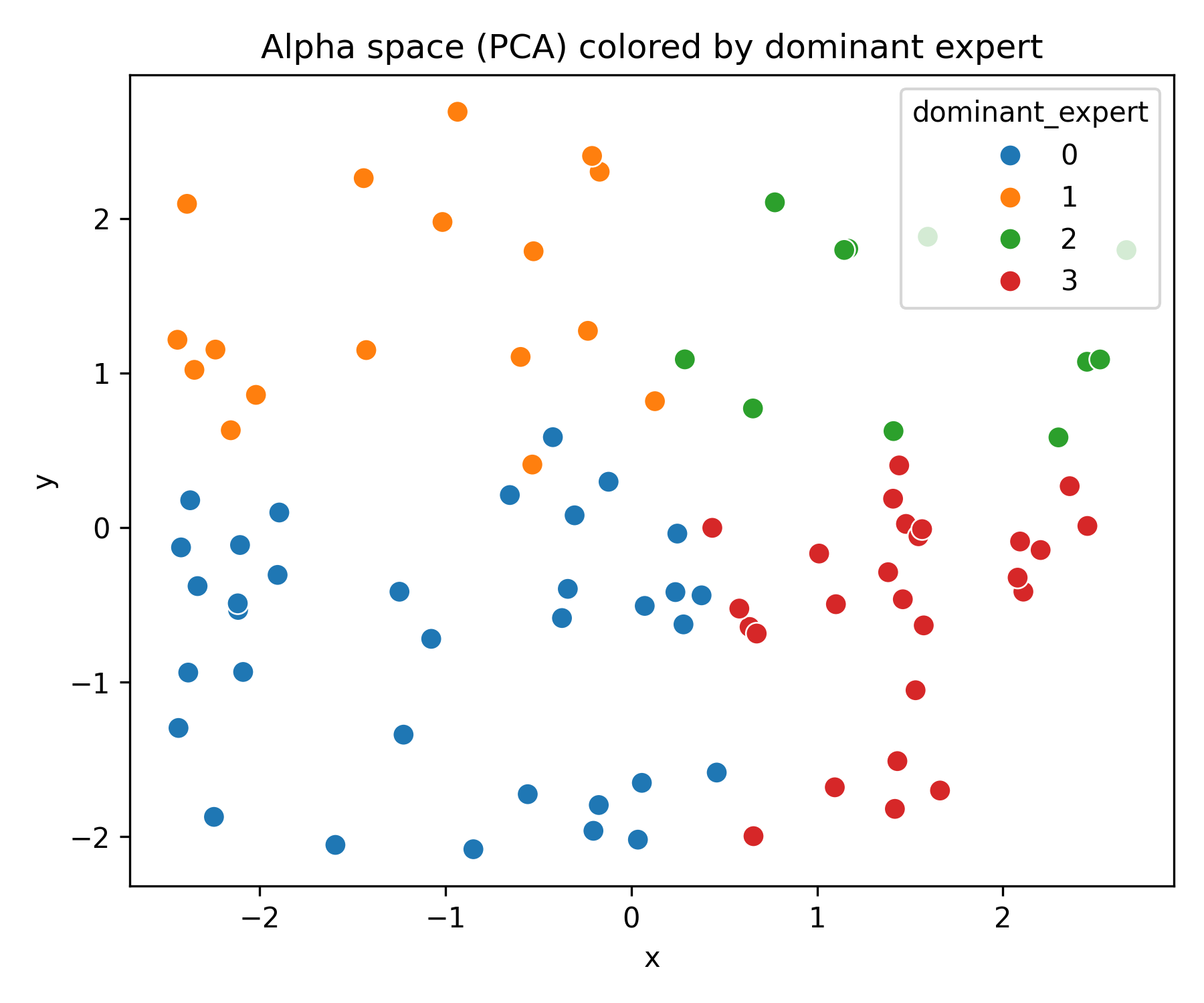}
    \caption{
    PCA projection of routing distributions ($\boldsymbol{\alpha}$),
    colored by dominant expert.
    The separation indicates that expert routing occupies
    distinct regions in routing space.
    }
    \label{fig:appendix-alpha-pca}
\end{figure}

\begin{figure*}[t]
    \centering
    \includegraphics[width=0.30\linewidth]{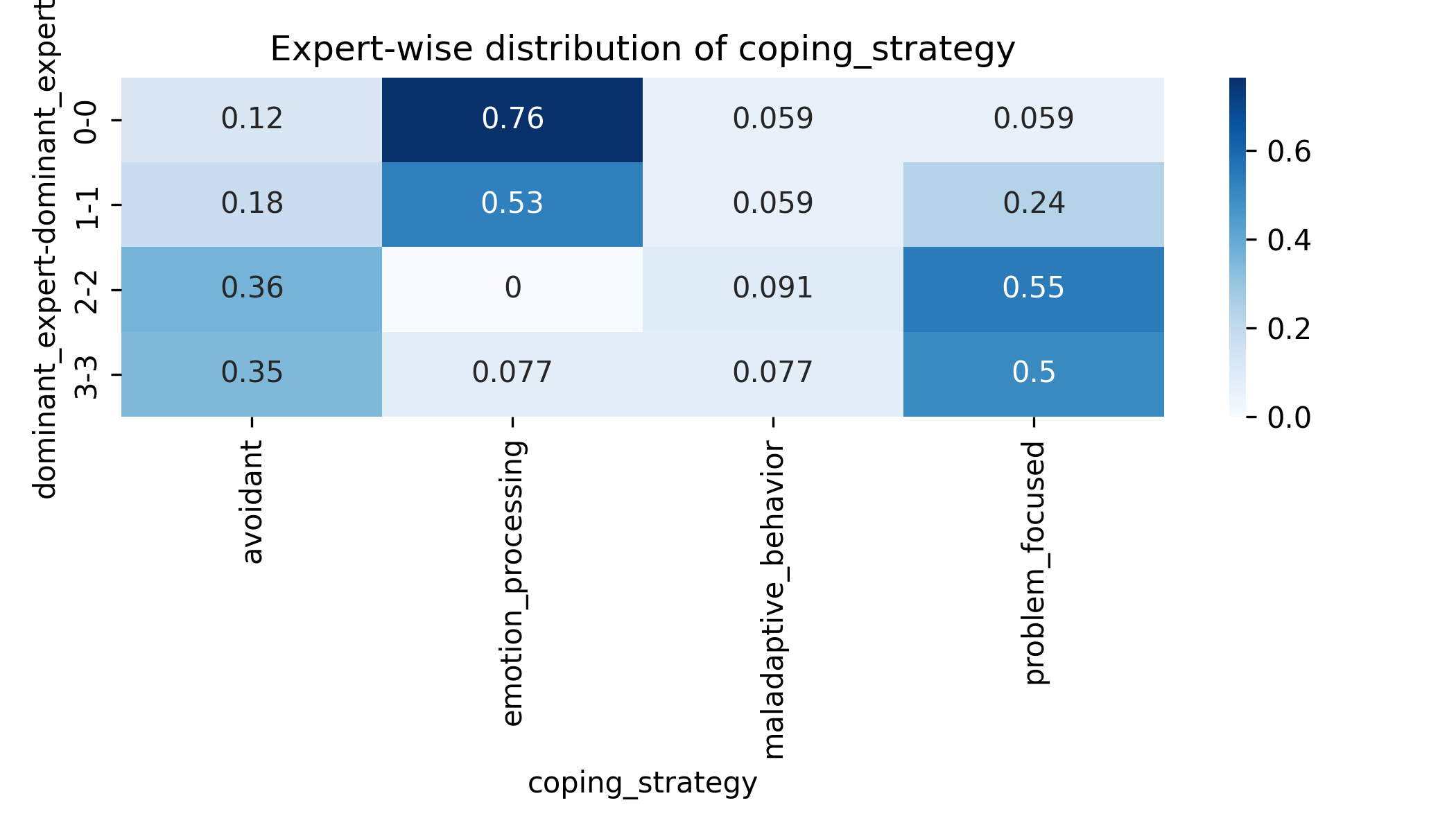}
    \includegraphics[width=0.30\linewidth]{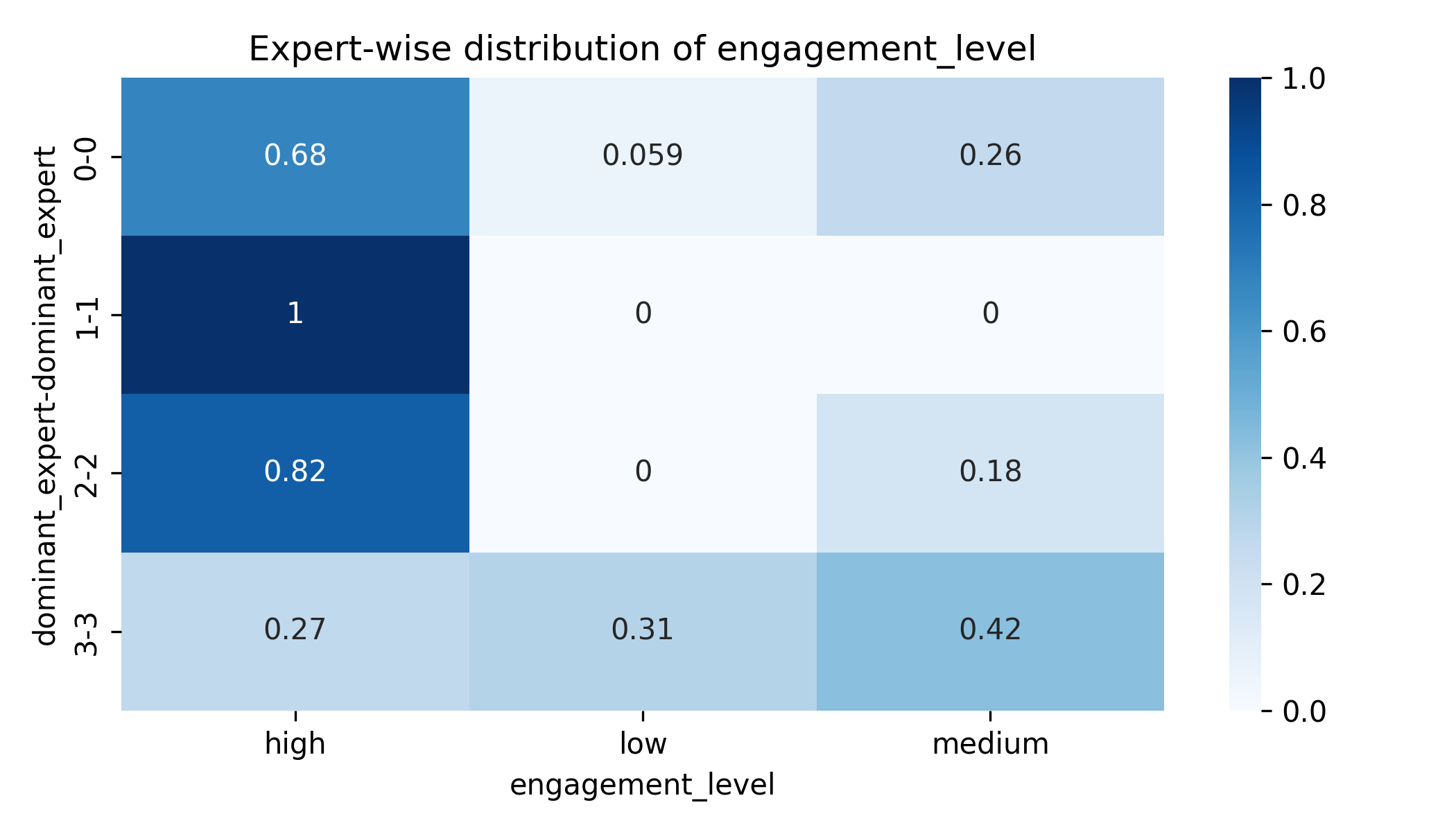}
    \includegraphics[width=0.30\linewidth]{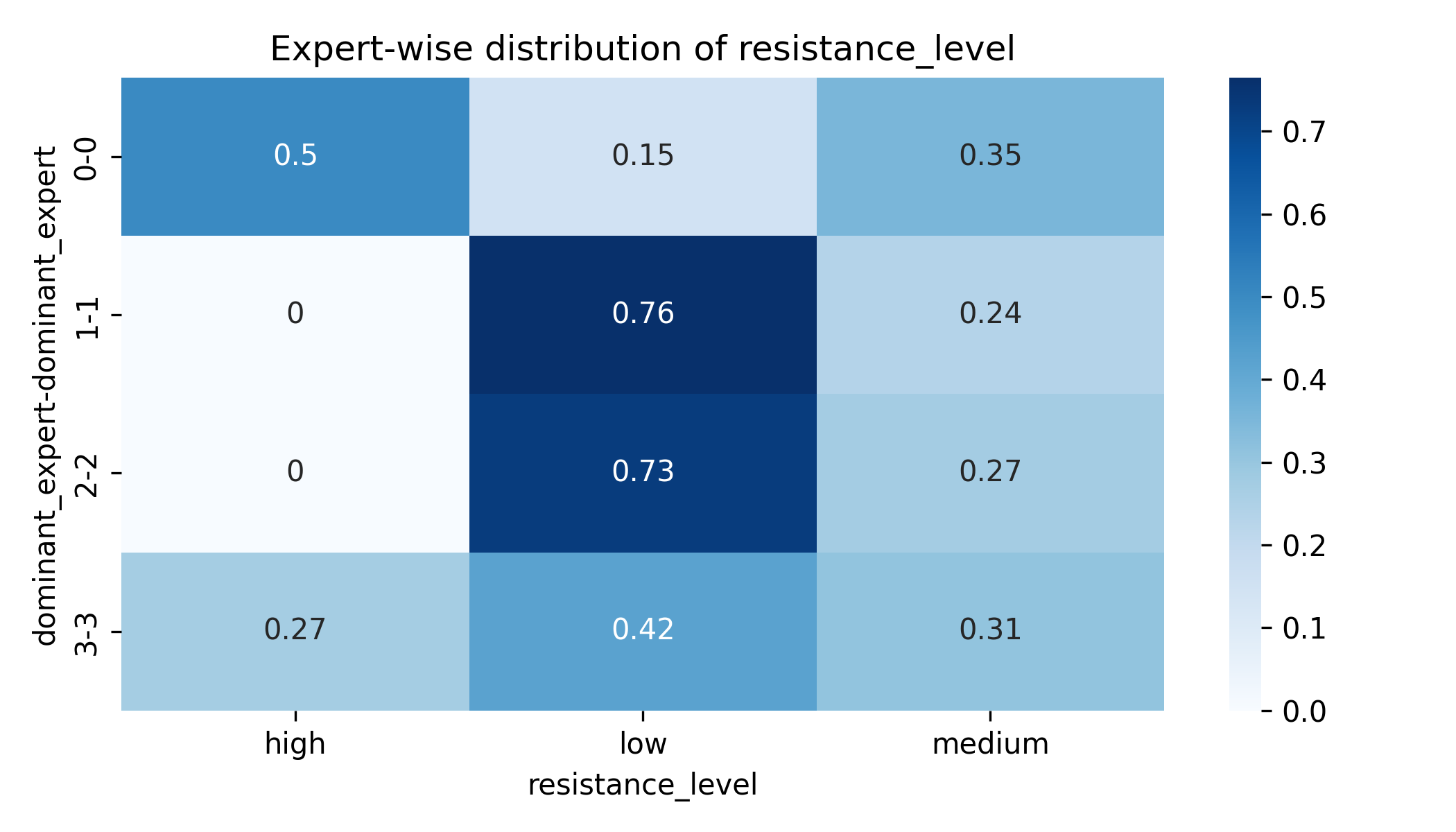}

    \vspace{0.5em}

    \includegraphics[width=0.30\linewidth]{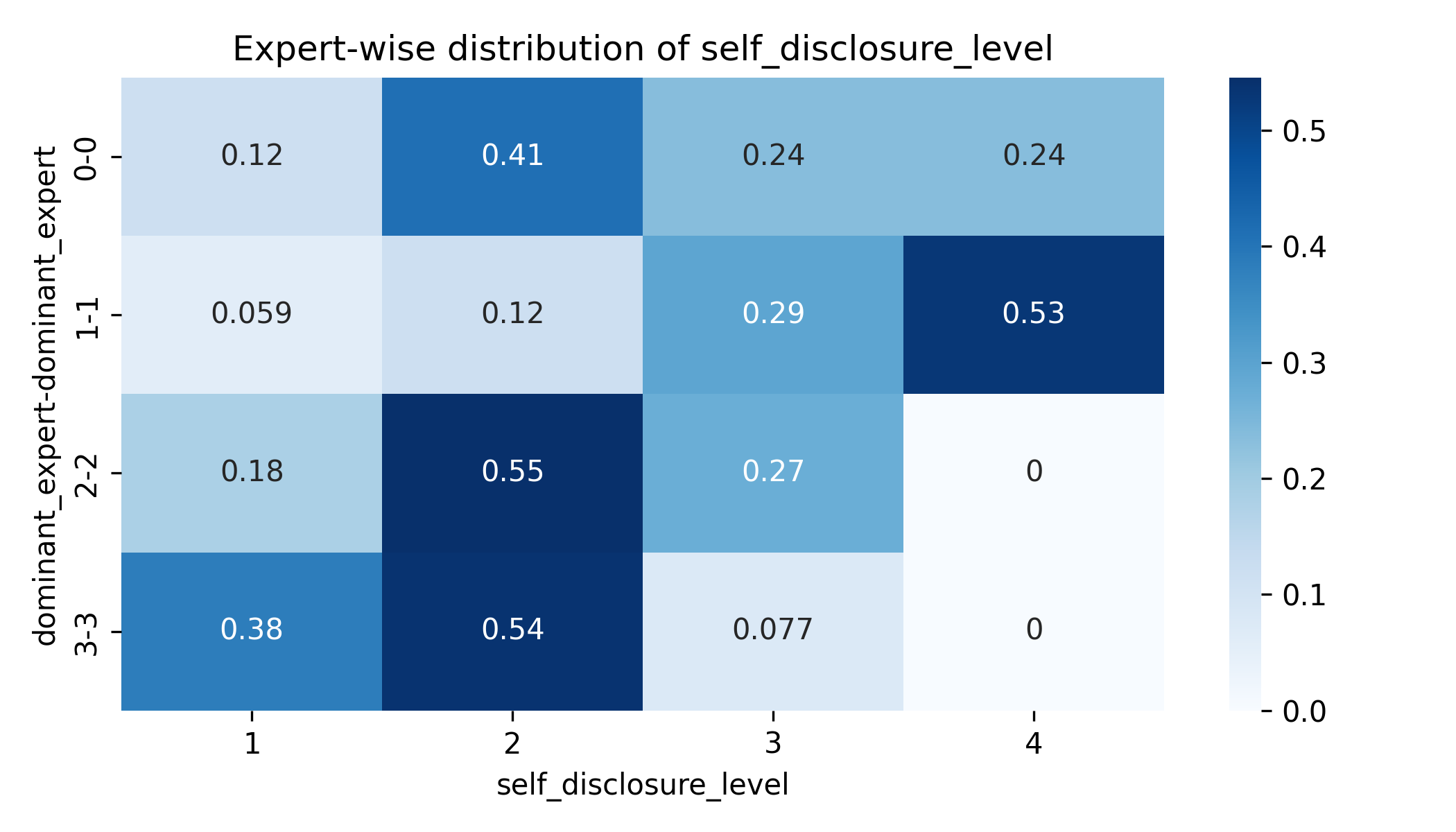}
    \includegraphics[width=0.30\linewidth]{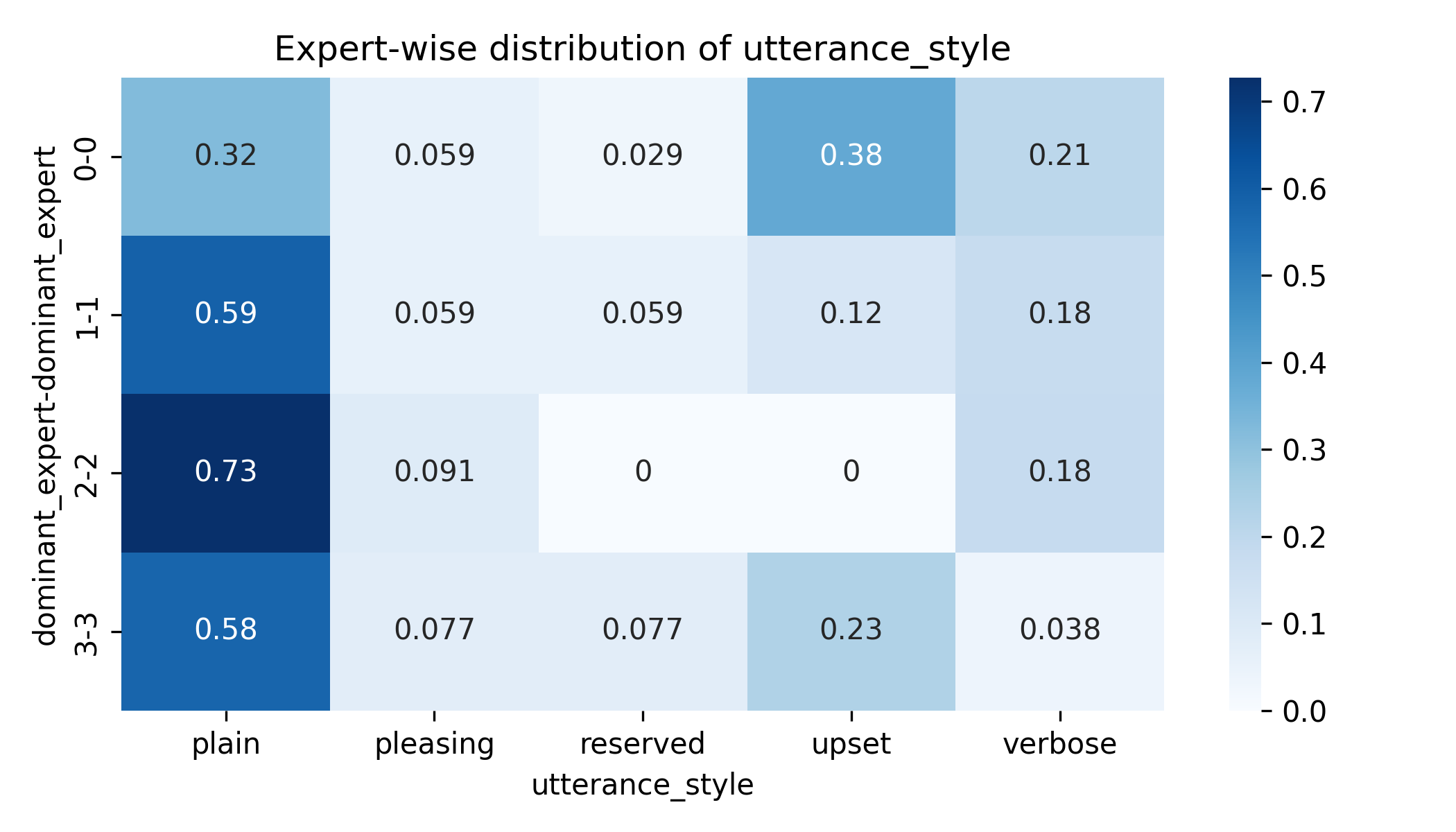}
    \caption{
    Each heatmap row is normalized to sum to 1, highlighting expert preference
    under each feature label.
    }
    \label{fig:appendix-expert-heatmaps}
\end{figure*}

Figure~\ref{fig:appendix-alpha-pca} visualizes the routing space via PCA. Each point corresponds to a dialogue sample and is colored by its dominant expert $e^* = \arg\max \boldsymbol{\alpha}$.
Distinct clustering by dominant expert suggests that the routing network learns separable regions in $\boldsymbol{\alpha}$ space, reinforcing the observed feature-conditioned specialization. This structural separation indicates that routing decisions are organized in a meaningful latent space, where each expert occupies a consistent region, supporting expert-level interpretation.
Figure~\ref{fig:appendix-expert-heatmaps} presents conditional distributions $P(\text{feature} \mid e^*)$ of behavioral features given the dominant expert. Each row is normalized to sum to 1. High values in a row indicate that a particular feature level is strongly associated with that expert. These patterns support the claim that experts capture distinct behavioral regimes rather than surface-level variations.

These distributions reveal distinct behavioral tendencies for each expert. For example, Expert 1 shows strong concentration on high engagement (1.0), low resistance (0.76), and high self-disclosure (level 4: 0.53), whereas Expert 3 exhibits relatively higher proportions of low engagement and lower self-disclosure levels (1–2). Such patterns demonstrate that expert routing aligns with coherent behavioral profiles rather than superficial variation.

Despite this expert-based structure, the MoE framework incurs minimal overhead. The routing network adds only 15,881 parameters (${\sim}$0.0003\% of total model size), requiring approximately 2 hours and 40 minutes of additional training on a single H100 GPU. Moreover, MoE provides targeted improvements on features that SFT largely fails to control, such as self-disclosure level (0.40 $\rightarrow$ 0.45) and resistance level (0.37 $\rightarrow$ 0.43), effectively expanding the range of reliably controllable behavioral dimensions.

\section{Expert Evaluations on Fidelity}
\label{app:expert-eval-fidelity}
\subsection{Metric Definitions}
\begin{itemize}
    \item Linguistic Naturalness: Does the seeker speak in a fluid, human-like manner without mechanical repetitions or awkward phrasing?
    \item Role Authenticity: Does the seeker maintain the role and behavioral traits consistent with a person seeking emotional support?
    \item Psychological Plausibility: Are the seeker’s emotional transitions and reactions psychologically coherent throughout the conversation?
\end{itemize}

\subsection{Expert Evaluation Settings}
\begin{figure*}[t]
  \centering
  \includegraphics[width=\textwidth]{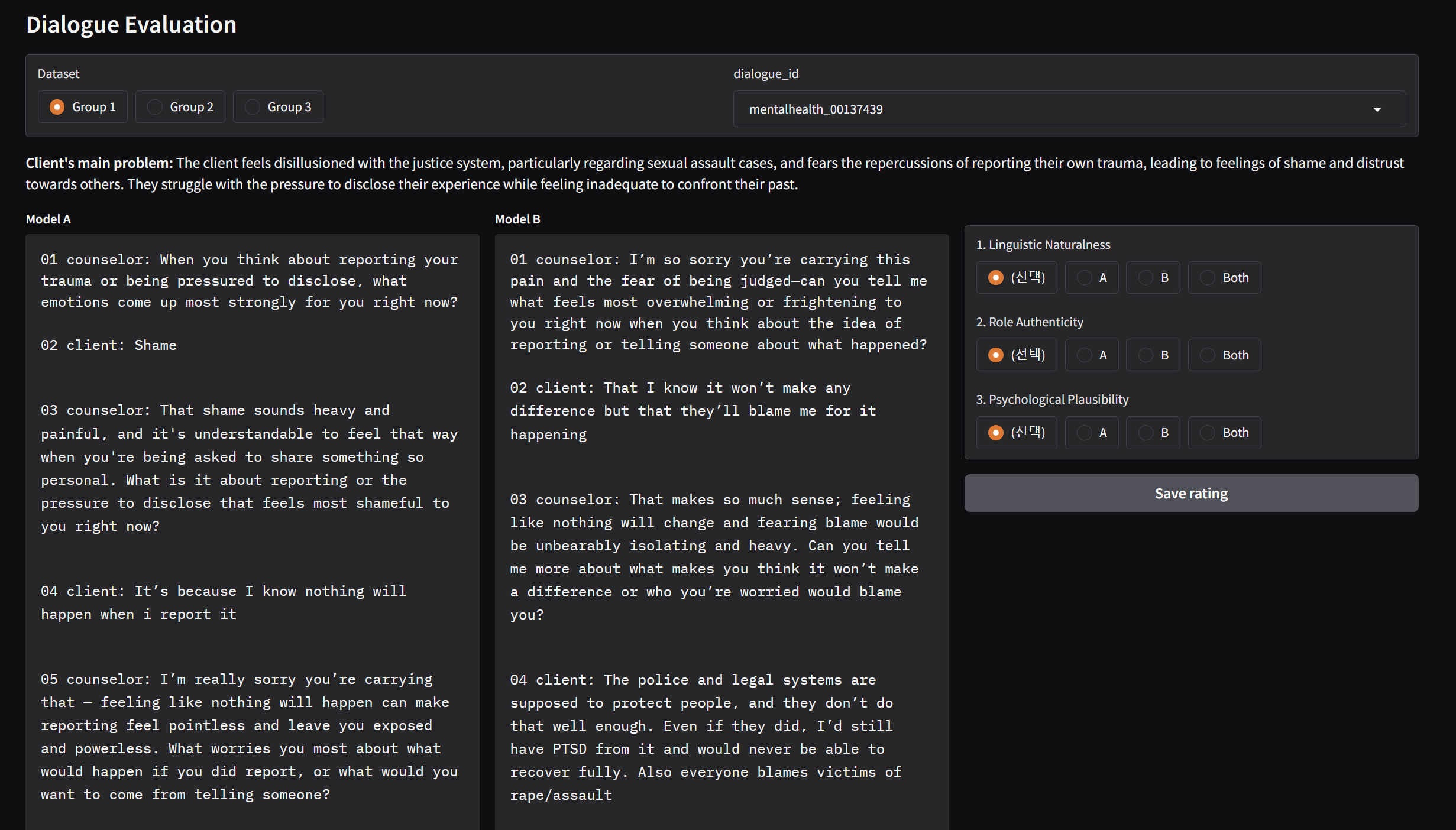}
  \caption{User interface for expert evaluation.}
  \label{fig:gradio-ui}
\end{figure*}

For expert evaluation, evaluators were first instructed on the criteria-specific guidelines, and then performed the assessment using the interface in Figure~\ref{fig:gradio-ui} based on the instructions in Table~\ref{tab:expert-eval-instruction}.

\paragraph{Annotator Details}
The evaluation was conducted by three graduate students specializing in clinical psychology (two males and one female). Participation was voluntary, and each annotator was compensated at approximately \$40 per hour in accordance with standard research participation guidelines.

\paragraph{Inter-Annotator Agreement}
We report inter-annotator agreement to assess the reliability of our expert evaluations. Annotators were asked to choose among ``prefer A'', ``prefer B'', and ``prefer both''. Given the ordinal nature of the scale (A~=~0, Both~=~0.5, B~=~1), we compute mean absolute ordinal deviation to capture partial agreement rather than treating all disagreements as equally distant. The resulting scores were 0.341, 0.315, and 0.322 for Linguistic Naturalness, Role Authenticity, and Psychological Plausibility, respectively, indicating that annotators rarely showed completely opposite preferences. Pairwise agreement and full agreement rates are reported in Table~\ref{tab:iaa}.

\begin{table}[h]
\centering
\small
\setlength{\tabcolsep}{5pt}
\renewcommand{\arraystretch}{1.1}
\begin{tabular}{l|cccc}
\toprule
\textbf{Metric} & \textbf{1--2} & \textbf{1--3} & \textbf{2--3} & \textbf{All} \\
\midrule
Linguistic Naturalness     & 0.811 & 0.433 & 0.411 & 0.367 \\
Role Authenticity          & 0.744 & 0.411 & 0.389 & 0.311 \\
Psychological Plausibility & 0.800 & 0.300 & 0.311 & 0.267 \\
\bottomrule
\end{tabular}
\caption{Pairwise and full inter-annotator agreement rates. Columns 1--2, 1--3, 2--3 denote pairwise agreement between annotators, and All denotes the rate of full agreement across all three annotators.}
\label{tab:iaa}
\end{table}

\section{Profile construction for each Seeker Simulator}
\label{app:profile-construction-each}
To ensure fair comparison and preserve the originally reported performance of each seeker simulator, we adopt the original profile structures and prompting schemes proposed in their respective papers without modification. We construct 300 held-out Reddit-derived profiles for each simulator following its originally proposed profile construction procedure. For ClientCAST, we instead use its originally proposed datasets directly, as the simulator requires reference dialogues to be provided during generation. In addition, when constructing simulator-specific profiles, we aim to cover a broad range of seeker characteristics within each simulator’s representational capacity by assigning diverse feature labels where applicable.

\paragraph{ESC-Judge}
For ESC-Judge, we construct seeker profiles following the original profile generation framework proposed in the paper. Each profile is generated by sampling an ongoing challenge from a predefined stressor set and composing a structured persona that includes demographic attributes, family context, and occupational background. Additional contextual information, such as past life experiences and behavioral traits is sampled from predefined spaces and integrated into a unified profile card.

\paragraph{ESC-Eval}
For ESC-Eval, we directly use the publicly released test profiles provided by the original work. To maintain diversity while matching the evaluation scale of other simulators, we randomly sample 300 profiles from the released test set, preserving the original distribution of profile attributes.

\paragraph{Eeyore}
For Eeyore, the number of publicly available test profiles is smaller than the required scale. To address this, we construct additional profiles following the procedure described in the original paper. Specifically, we extract each seeker’s main problem from the training dataset using the same extraction method employed in our Reddit data preprocessing pipeline and assign it to the client situation category. We then assign categorical labels to promote diversity across feature dimensions, such as varying symptom severity, balancing resistance toward support, and diversifying persona attributes, while maintaining balanced coverage across feature values.

\paragraph{ClientCAST}
For ClientCAST, we reproduce the simulator following the methodology and codebase described in the original paper. Seeker profiles are constructed from two counseling datasets: the High-Low Quality Counseling dataset~\cite{perez-rosas-etal-2019-makes} and AnnoMI~\cite{9746035}, from which we extract 213 high- and 87 low-quality sessions as specified in the original work. As ClientCAST requires reference dialogues during simulation, we directly use the original counseling sessions from these datasets. Each seeker profile consists of three components: (1) Problems \& Reasons for Visiting, (2) Displayed Symptoms from a predefined set of 61 client symptoms, and (3) Apparent Traits, assigned to promote diversity across seeker characteristics.

\section{Diversity Metrics}
\label{app:diversity-metrics}
We employ embedding-space visualizations together with lexical, semantic, and sentiment-based metrics to capture the diversity of the simulators’ outputs.

\subsection{Visualization Metrics}
\paragraph{UMAP-based Visualization Metric} We visualize seeker diversity by projecting dialogue-level embeddings into a two-dimensional space using UMAP (Uniform Manifold Approximation and Projection). Each dialogue is represented by a single embedding obtained by aggregating all seeker utterances and encoding them with a sentence embedding model. UMAP is applied to the combined embeddings from all seeker simulators using cosine distance, preserving local neighborhood structure while maintaining coarse global relationships in the low-dimensional space.

\paragraph{Convex Hull-based Coverage Metrics}
To complement the qualitative patterns observed in the UMAP visualizations, we additionally quantify the spatial coverage of each seeker simulator using convex hull area metrics. Specifically, we compute the area of the convex hull enclosing dialogue-level embeddings projected onto two-dimensional spaces obtained via t-SNE and PCA. These metrics provide a simple geometric estimate of how broadly each simulator’s generated dialogues are distributed in the embedding space, serving as a quantitative proxy for the visual spread observed in the projections.

Table~\ref{tab:diversity-hull} reports the convex hull areas for each simulator. Consistent with the UMAP visualizations, our simulator exhibits the largest hull area under both t-SNE and PCA projections, indicating broader coverage of the embedding space compared to baseline simulators. In contrast, baseline simulators occupy more confined regions, reflecting more limited diversity in generated seeker expressions.

\begin{table}[t]
\centering
\small
\setlength{\tabcolsep}{2.5pt}
\renewcommand{\arraystretch}{1.1}
\begin{tabular}{lcc}
\toprule
\textbf{Seeker Simulator} & \textbf{TSNE Hull Area}~$\uparrow$ & \textbf{PCA Hull Area}~$\uparrow$ \\
\midrule
ClientCAST & 2954.562 & 0.312 \\
ESC-Judge  & 1878.463 & 0.334 \\
ESC-Role   & 3093.516 & 0.378 \\
Eeyore     & 3057.489 & 0.441 \\
\textbf{Ours} & \textbf{3378.139} & \textbf{0.659} \\
\bottomrule
\end{tabular}
\caption{Convex hull areas computed on t-SNE and PCA projections of dialogue-level seeker embeddings, reflecting the spatial coverage of each simulator.}
\label{tab:diversity-hull}
\end{table}

\subsection{Quantitative Metrics}
In addition to the visualization-based analysis described above, we employ a set of quantitative diversity metrics to measure lexical variation, semantic dispersion, and sentiment distribution in a complementary and reproducible manner. These metrics provide numerical estimates of diversity that support the qualitative patterns observed in the embedding-space visualizations.

\paragraph{Metric Definitions}
\begin{itemize}
\item \textbf{Lexical Diversity}  
We measure lexical diversity using Distinct-$n$, Self-BLEU, and Token Repetition Mean, computed over dialogue-level seeker texts.
Distinct-$n$ is calculated as the ratio of unique $n$-grams to the total number of $n$-grams across all dialogues, capturing surface-level vocabulary variety.
Self-BLEU is computed by treating each dialogue as a hypothesis and the remaining dialogues as references, and averaging BLEU scores across dialogues to quantify intra-set redundancy.
Token Repetition Mean is calculated as the average proportion of repeated tokens within each dialogue, reflecting repetitive word usage.

\item \textbf{Semantic Diversity}  
To assess semantic diversity, we compute Average Pairwise Distance (APD) and Average Cosine Similarity (ACS) over dialogue-level embedding representations.
Each dialogue is embedded by encoding the concatenation of all seeker utterances using a sentence embedding model.
APD is computed as the mean cosine distance over all unique pairs of dialogue embeddings, measuring the overall semantic dispersion of generated dialogues.
ACS is computed as the mean cosine similarity over the same embedding pairs, capturing the degree of semantic redundancy.
Higher APD (and correspondingly lower ACS) indicates greater semantic diversity among dialogue representations.

\item \textbf{Sentiment Diversity}  
We evaluate sentiment diversity using VADER compound sentiment scores computed for each dialogue.
For each simulator, we report the mean and variance of compound scores across dialogues.
The mean reflects the overall sentiment tendency, while the variance captures the spread of expressed emotional polarity across generated seeker utterances.

\end{itemize}

\paragraph{Results and Analysis}
Across all lexical and semantic diversity metrics, our simulator consistently achieves the highest diversity scores.

Specifically, it attains the strongest performance on lexical diversity measures including Distinct-2, Self-BLEU, and Token Repetition Mean, indicating richer and less repetitive vocabulary usage.
Similarly, on semantic diversity metrics—Average Pairwise Distance (APD) and Average Cosine Similarity (ACS)—our simulator exhibits the largest semantic dispersion and the lowest redundancy among dialogue representations.

We further analyze affective diversity using VADER compound scores. Our simulator yields a sentiment mean of $-0.05$ with a variance of $0.70$, reflecting a wide distribution of emotional states spanning from negative to positive affect. In contrast, baseline simulators exhibit sentiment means above $0.7$ with variances below $0.5$, suggesting a strong bias toward uniformly positive expressions and limited emotional variability. These results indicate that existing simulators tend to generate overly optimistic seeker behaviors, whereas our model more faithfully captures the emotional heterogeneity observed in real-world emotional support seekers, including both negative and positive affective states.

\section{Evaluation Settings for Supporter Model Evaluation}
\label{app:evaluation-details-for-supporter-eval}
\paragraph{Evaluation Settings}
To ensure comparable initial conditions across evaluations, dialogues are initialized using seeker utterances derived from the seeker’s main problem. This design choice standardizes the starting context and prevents variations in supporter performance caused by inconsistent dialogue entry points.

All seeker simulators are evaluated using held-out profiles that preserve the original profile structures and prompting schemes specified in prior work. This strategy maintains the intended operating conditions of each simulator while enabling controlled comparison across different seeker populations.

Dialogue termination is implemented differently depending on the simulator. In standard settings, conversations naturally conclude through mutual closing exchanges, such as expressions of gratitude or farewell. In contrast, our seeker simulator explicitly models dialogue termination by conditioning generation on the total dialogue turns level feature and emitting a dedicated \texttt{<|end\_of\_dialogue|>} token. This mechanism is learned from training data and allows the simulator to regulate dialogue length in a principled and interpretable manner.

The maximum dialogue length constraint is informed by empirical statistics of the underlying data, where seeker–supporter interactions exhibit an average length of approximately 17 turns. This observation motivates the dialogue length cap used in evaluation.

\paragraph{Prompts for Supporter Models}
To ensure fairness across supporter models, a unified supporter role prompt is applied throughout all evaluations. The full prompt template is provided in Figure~\ref{fig:supporter-prompt}.

\begin{figure}[htbp]  
\centering  
\begin{tcolorbox}[
  colback=gray!5,
  colframe=gray!50,
  boxrule=0.6pt,
  arc=3pt,
  left=6pt,
  right=6pt,
  top=6pt,
  bottom=6pt,
  width=\columnwidth  
]
\textbf{Supporter role prompt}

\smallskip
\raggedright
\small
\ttfamily
You are a supportive and empathetic assistant for an emotional support context. Respond in a calm and non-judgmental manner, acknowledging and validating the user's emotions.\\
Ask gentle clarifying questions only when helpful, and offer practical coping suggestions when appropriate.\\
Do not diagnose, prescribe, or claim professional authority.\\
Keep responses concise and warm (3--6 sentences).
\end{tcolorbox}
\caption{Supporter role prompt template for the emotional support simulator.}
\label{fig:supporter-prompt}  
\end{figure}

\section{Datasets Used for Fine-tuning Supporter Models}
\label{app:supporter-train-datasets}
The supporter models were fine-tuned on publicly available emotional support datasets,
including ESConv~\cite{liu-etal-2021-towards}, ExTES~\cite{zheng-etal-2024-self},
Psych8k~\cite{liu2023chatcounselorlargelanguagemodels}, PsyInsight~\cite{chen2025psyinsightexplainablemultiturnbilingual}, and CounselChat~\cite{bertagnolli2020counselchat}.

\section{Human Evaluation Details}
\label{app:human_eval_details}

\paragraph{Experimental Setup}
We selected Ours and ESC-Judge as the two seeker simulators, as they produced the largest performance gaps in supporter evaluation. For supporter models, we included GPT-5-mini and Llama-ESConv, representing prompt-based and fine-tuned supporters, respectively. From the dialogues generated between these seeker simulators and supporter models, we sampled 60 dialogues ensuring a uniform distribution of score discrepancies, covering the full spectrum from cases with similar scores to those with substantial differences. Beyond the $\Delta$-based analysis presented in Section~\ref{sec:automatic_evaluation_validity}, we additionally report Human--LLM score correlations and inter-annotator agreement to verify the reliability of the automated evaluation and the human evaluation itself, respectively. 

\paragraph{Human--LLM Raw Score Correlations}
To assess whether the automated scoring aligns with human judgments at the absolute score level, we computed Pearson and Spearman correlations between human and automated scores. The resulting correlations were moderate to strong across all ES-skills (all $p < 0.001$), indicating that the automated evaluation reliably reflects human assessments. (Table~\ref{tab:human_llm_corr})

\begin{table}[h]
\centering
\small
\setlength{\tabcolsep}{5pt}
\renewcommand{\arraystretch}{1.1}
\begin{tabular}{l|cc}
\toprule
\textbf{Metric} & \textbf{Pearson $r$} & \textbf{Spearman $\rho$} \\
\midrule
Identification   & 0.650 & 0.628 \\
Comforting       & 0.589 & 0.546 \\
Suggestions      & 0.730 & 0.761 \\
Exp. Sharing     & 0.636 & 0.541 \\
Informativeness  & 0.807 & 0.794 \\
\bottomrule
\end{tabular}
\caption{Pearson and Spearman correlations between human and automated evaluation scores.}
\label{tab:human_llm_corr}
\end{table}

\paragraph{Inter-Annotator Agreement}
To verify the reliability of the human evaluation itself, we assessed inter-annotator agreement using quadratic weighted Cohen's $\kappa$, which penalizes larger disagreements more heavily. Agreement ranged from fair to strong across metrics, indicating generally reliable human evaluation. (Table~\ref{tab:iaa_human_eval})

\begin{table}[h]
\centering
\small
\setlength{\tabcolsep}{5pt}
\renewcommand{\arraystretch}{1.1}
\begin{tabular}{l|c}
\toprule
\textbf{Metric} & \textbf{Quadratic $\kappa$} \\
\midrule
Identification   & 0.595 \\
Comforting       & 0.345 \\
Suggestions      & 0.658 \\
Exp. Sharing     & 0.821 \\
Informativeness  & 0.774 \\
\bottomrule
\end{tabular}
\caption{Inter-annotator agreement measured by quadratic weighted Cohen's $\kappa$.}
\label{tab:iaa_human_eval}
\end{table}

\begin{table*}[t]
\centering
\footnotesize
\begin{tabular}{p{0.18\linewidth} p{0.76\linewidth}}
\toprule
\textbf{Feature} & \textbf{LLM Tagging Prompt} \\
\midrule

\textit{Main Coping Strategy} &
\textbf{Task.} Evaluate the client’s main coping strategy, defined as the single most dominant strategy the client uses across the entire dialogue to manage stress, based only on the client’s utterances. \newline
\textbf{Instructions.}
(1) Read the entire dialogue. Read the client’s main problem for context only.
(2) Base the evaluation only on the client’s utterances.
(3) Assign a single coping strategy label that best represents the client’s dominant and consistent approach. \newline
\textbf{Labels.}
\textit{problem\_focused}: concrete actions or cognitive efforts to address the stressor;
\textit{emotion\_processing}: emotional expression or processing without problem solving;
\textit{avoidant}: deliberate avoidance of the problem or emotions;
\textit{maladaptive\_behavior}: current self-destructive coping behavior. \\

\addlinespace
\textit{Engagement Level} &
\textbf{Task.} Evaluate the client’s overall engagement level throughout the entire dialogue, reflecting how active, serious, and honest the client is in the interaction. \newline
\textbf{Instructions.}
(1) Read the entire dialogue. Read the client’s main problem for context only.
(2) Base the evaluation only on the client’s utterances.
(3) Assign a single engagement level representing the client’s overall mode of participation. \newline
\textbf{Labels.}
\textit{high}: active and reflective participation;
\textit{medium}: inconsistent or biased engagement;
\textit{low}: minimal or disengaged participation. \\

\addlinespace
\textit{Resistance Level} &
\textbf{Task.} Evaluate the overall intensity and consistency of resistance behaviors exhibited by the client across the entire dialogue. \newline
\textbf{Instructions.}
(1) Read the entire dialogue. Read the client’s main problem for context only.
(2) Base the evaluation only on the client’s utterances.
(3) Assign a single resistance level reflecting the client’s dominant behavioral pattern. \newline
\textbf{Labels.}
\textit{high}: frequent and pervasive resistance;
\textit{medium}: situational resistance triggered by specific topics or interventions;
\textit{low}: generally cooperative with little or no resistance. \\

\addlinespace
\textit{Utterance Style} &
\textbf{Task.} Evaluate the client’s dominant utterance style used throughout the dialogue, focusing on how the client speaks rather than what is said. \newline
\textbf{Instructions.}
(1) Read the entire dialogue. Read the client’s main problem for context only.
(2) Base the evaluation only on the client’s utterances.
(3) Assign a single utterance style that best characterizes the client’s speaking pattern. \newline
\textbf{Labels.}
\textit{plain}: direct, clear, and to-the-point communication without excessive emotion or avoidance;
\textit{upset}: expression of frustration, anger, or strong resistance in a confrontational or dismissive tone;
\textit{verbose}: excessively long and detailed utterances that disrupt conversational flow;
\textit{reserved}: minimal, vague, or evasive responses requiring repeated prompting;
\textit{tangent}: deviation from the main topic into unrelated content;
\textit{pleasing}: a persistent pattern of excessive agreement or problem minimization toward the counselor. \\

\bottomrule
\end{tabular}
\caption{LLM prompts used for dialogue-level feature tagging.}
\label{tab:dialogue_level_raw_prompts}
\end{table*}

\begin{table*}[t]
\centering
\footnotesize
\begin{tabular}{p{0.18\linewidth} p{0.76\linewidth}}
\toprule
\textbf{Feature} & \textbf{LLM Tagging Prompt} \\
\midrule

\textit{Seeker Reaction} &
\textbf{Task.} Evaluate how the seeker reacts to the counselor’s immediately preceding utterance, focusing only on the seeker’s response rather than the seeker’s general emotional state or situation. \newline
\textbf{Instructions.}
(1) Read the client’s main problem for context only.
(2) Read the counselor utterance and the immediately following seeker utterance.
(3) Base the evaluation only on the seeker’s utterance.
(4) Assign a single reaction label that best represents the seeker’s response to the counselor. \newline
\textbf{Labels.}
\textit{positive}: openness, cooperation, acknowledgment, agreement, appreciation, or constructive engagement with the counselor’s message;
\textit{neutral}: vague, minimal, or non-committal responses without clear acceptance or rejection;
\textit{negative}: rejection, resistance, defensiveness, deflection, or pushback toward the counselor’s message. \\

\addlinespace
\textit{Self-Disclosure Level} &
\textbf{Task.} Evaluate the depth of personal information or emotional vulnerability revealed in the seeker’s utterance in response to the counselor’s immediately preceding message. \newline
\textbf{Instructions.}
(1) Read the client’s main problem for context only.
(2) Read the counselor utterance and the immediately following seeker utterance.
(3) Base the evaluation only on the seeker’s utterance.
(4) Assign a single self-disclosure level reflecting the depth of revealed information. \newline
\textbf{Labels.}
\textit{1 (orientation)}: surface-level, socially normative information without emotional content or personal meaning;
\textit{2 (exploratory affective exchange)}: personal facts or light emotions without deep vulnerability;
\textit{3 (affective exchange)}: emotionally meaningful experiences, fears, or vulnerabilities with personal significance;
\textit{4 (stable exchange)}: identity-level beliefs, core assumptions, or existential conclusions that generalize beyond specific events. \\

\bottomrule
\end{tabular}
\caption{LLM prompts used for utterance-level feature tagging.}
\label{tab:turn_level_raw_prompts}
\end{table*}


\begin{figure*}[t]
\centering
\begin{tcolorbox}[
  colback=gray!5,
  colframe=gray!50,
  boxrule=0.6pt,
  arc=3pt,
  left=6pt,
  right=6pt,
  top=6pt,
  bottom=6pt,
  width=\textwidth
]

\texttt{<seeker\_features>}\\
coping\_strategy: emotion\_processing \\
resistance\_level: high \\
engagement\_level: high \\
self\_disclosure\_level: 2 \\
seeker\_reaction: \{"positive\_ratio": 0.0, "neutral\_ratio": 0.0, "negative\_ratio": 1.0\} \\
utterance\_style: upset \\
verbosity\_level: medium (30--59 tokens) \\
user\_profanity\_flag: True \\
total\_dialogue\_turns\_level: long ($\ge$ 9 turns)

\medskip
\texttt{<seeker's\_main\_problem>}\\
The seeker is struggling with feelings of failure after not being accepted into medical school, despite having strong academic credentials. This obsession with medicine overshadows their ability to focus on job searching or other career paths, leading to persistent emotional distress and self-doubt.

\end{tcolorbox}
\caption{An example of seeker profile.}
\label{fig:seeker_profile_example}
\end{figure*}


\begin{table*}[t]
\centering
\small
\setlength{\tabcolsep}{6pt}
\renewcommand{\arraystretch}{1.1}
\begin{tabularx}{\textwidth}{p{0.23\textwidth} X}
\toprule
\textbf{System Prompt} &
You are a summarization assistant. \\
\midrule
\textbf{User Prompt} &
You are an evaluator. You will be provided with a first-person Reddit post written by a seeker. Your task is to write a concise summary that captures the seeker's situation and main problem.

\vspace{0.6em}
Please follow these steps:

\vspace{0.4em}
1. Read the post carefully and extract the context (what happened) and the core concern or emotional struggle.

\vspace{0.3em}
2. Write the summary in the third person (do not use ``I'', ``my'', etc).

\vspace{0.3em}
3. The summary must begin with ``problem:''

\vspace{0.3em}
4. Keep the summary to 1--2 sentences.

\vspace{0.3em}
5. Do not include any information that is not clearly stated in the post.

\vspace{0.3em}
6. Do not include any names or proper nouns. You may mention age or gender only when it is explicitly mentioned in the post, and never guess or infer it.

\vspace{0.6em}
[Post]  

\{OP text\}

\vspace{0.6em}
[Output format]  

problem: [summary in third person, 1--2 sentences, capturing the situation and the seeker's emotional or psychological struggle] \\
\hline
\end{tabularx}
\caption{Prompt template for extracting seeker's main problem from first-person Reddit posts.}
\label{tab:seeker-summary-prompt}
\end{table*}


\clearpage

\begin{table*}[t]
\centering
\captionsetup{justification=centering}

\small
\setlength{\tabcolsep}{4pt}
\renewcommand{\arraystretch}{1.1}
\begin{tabular}{l|cccccccc}
\toprule
\textbf{Model} 
& \textbf{Coping} 
& \textbf{Utterance} 
& \textbf{Engagement} 
& \textbf{Verbosity} 
& \textbf{Self-Disclosure} 
& \textbf{Resistance} 
& \textbf{Profanity} 
& \textbf{Turns} \\
\midrule
GPT-4.1-mini              & 0.33 & 0.36 & 0.27 & 0.28 & 0.19 & 0.24 & 0.58 & 0.16 \\
Llama-3-8B-Instruct       & 0.29 & 0.24 & 0.27 & 0.11 & 0.14 & 0.29 & 0.58 & 0.15 \\
Qwen-2.5-14B-Instruct     & 0.25 & 0.32 & 0.29 & 0.33 & 0.22 & 0.24 & 0.47 & 0.15 \\
GPT-5                     & 0.23 & 0.33 & 0.27 & 0.23 & 0.25 & 0.25 & 0.84 & 0.15 \\
DeepSeek-V3.2             & 0.39 & 0.40 & 0.41 & 0.52 & 0.27 & 0.37 & \textbf{0.91} & 0.18 \\
SFT                       & 0.38 & 0.36 & 0.53 & 0.66 & 0.40 & 0.37 & 0.66 & 0.56 \\
Contrastive Learning      & 0.30 & 0.32 & 0.39 & 0.54 & 0.32 & 0.35 & 0.64 & \textbf{0.83} \\
\textbf{Ours}             & \textbf{0.44} & \textbf{0.47} & \textbf{0.58} & \textbf{0.69} & \textbf{0.45} & \textbf{0.43} & 0.63 & 0.74 \\
\bottomrule
\end{tabular}
\caption{Feature-wise Macro F1 Scores.}
\label{tab:all_feature_macro_f1}

\vspace{8.0em}

\setlength{\tabcolsep}{6pt}
\renewcommand{\arraystretch}{1.15}
\begin{tabular}{lcccc}
\toprule
\textbf{Simulator}
& \textbf{Engagement}
& \textbf{Verbosity}
& \textbf{Self-Disclosure}
& \textbf{Resistance} \\
\midrule
GPT-4.1-mini               & --   & 0.81 & 0.42 & 0.15 \\
Llama-3-8B-Instruct        & 0.06 & 0.16 & 0.07 & 0.15 \\
Qwen-2.5-14B-Instruct      & 0.09 & 0.66 & 0.27 & 0.10 \\
GPT-5                      & --   & \textbf{0.89} & 0.42 & 0.20 \\
DeepSeek-V3.2              & 0.38 & 0.85 & 0.47 & 0.34 \\
SFT                        & 0.50 & 0.67 & 0.51 & 0.23 \\
Contrastive Learning       & 0.30 & 0.55 & 0.39 & 0.40 \\
\textbf{Ours}              & \textbf{0.52} & 0.72 & \textbf{0.58} & \textbf{0.42} \\
\bottomrule
\end{tabular}
\caption{Feature-wise Pearson correlations. Entries marked as ``--'' indicate correlations that could not be computed due to zero variance in one of the variables.}
\label{tab:feature_profile_correlations}
\end{table*}
\clearpage



\begin{table*}[t]
\centering
\small
\setlength{\tabcolsep}{8pt}
\renewcommand{\arraystretch}{1.6}

\begin{tabular}{p{0.22\textwidth} p{0.73\textwidth}}
\hline
\textbf{Metric} & \textbf{Evaluation Instruction} \\
\hline

\textbf{Linguistic Naturalness} &
\textbf{Definition:}
How much the client’s utterances sound like a real person in emotional support conversations, rather than language that appears overly refined or AI-generated. \newline
\textbf{Instruction:}
Assess whether the client’s language resembles natural, spontaneous human speech in emotional support contexts.
\\
\hline

\textbf{Role Authenticity} &
\textbf{Definition:}
How well the speaker stays in the role of a person seeking help, rather than sounding like a counselor, evaluator, or detached observer. \newline
\textbf{Instruction:}
Evaluate whether the client speaks from a first-person, emotionally grounded perspective, without excessive self-analysis or professional terminology.
\\
\hline

\textbf{Psychological Plausibility} &
\textbf{Definition:}
Whether the client’s emotional reactions and changes over the course of the dialogue feel realistic for a human in a difficult emotional situation. \newline
\textbf{Instruction:}
Examine whether emotional shifts occur gradually and proportionally in response to the conversation, rather than resolving abruptly.
\\
\hline

\end{tabular}
\caption{Expert evaluation instructions for comparing client utterances across counseling dialogues.}
\label{tab:expert-eval-instruction}
\end{table*}


\begin{table*}[t]
\centering
\small
\renewcommand{\arraystretch}{1.1}

\resizebox{0.9\linewidth}{!}{%
\begin{tabular}{l|ccc|ccc|ccc}
\hline
\multirow{2}{*}{\textbf{Comparison}}
& \multicolumn{3}{c|}{\textbf{Linguistic Naturalness}}
& \multicolumn{3}{c|}{\textbf{Role Authenticity}}
& \multicolumn{3}{c}{\textbf{Psychological Plausibility}} \\
\cline{2-10}
& Win & Loss & Tie & Win & Loss & Tie & Win & Loss & Tie \\
\hline
Ours vs. Eeyore     & 62 & 19 &  9 & 60 & 20 & 10 & 64 & 13 & 13 \\
Ours vs. ESC-Judge  & 62 & 18 & 10 & 65 & 19 &  6 & 56 & 14 & 20 \\
Ours vs. ESC-Role   & 72 &  9 &  9 & 61 & 16 & 13 & 61 & 12 & 17 \\
\hline
\end{tabular}%
}

\caption{Expert evaluation comparing our seeker simulator with existing baselines across three evaluation criteria. 
Win/Loss/Tie denote the number of instances where our simulator is preferred over, dispreferred to, or tied with the baseline.}
\label{tab:human-pref-metric}
\end{table*}


\begin{table*}[t]
\centering
\small
\setlength{\tabcolsep}{6pt}
\begin{tabular}{lccc|cc|cc}
\toprule
\textbf{Simulator} 
& \multicolumn{3}{c|}{\textbf{Lexical Diversity}} 
& \multicolumn{2}{c|}{\textbf{Semantic Diversity}} 
& \multicolumn{2}{c}{\textbf{Sentiment Diversity}} \\
 & Distinct-2~$\uparrow$ 
 & Self-BLEU~$\downarrow$ 
 & TokenRep~$\downarrow$ 
 & APD~$\uparrow$ 
 & ACS~$\downarrow$ 
 & SentMean 
 & SentStd \\
\midrule
ClientCAST
& 0.3031 & 0.8634 & 0.5329
& 0.5832 & 0.4168
& 0.9569 & 0.2619 \\
ESC-Judge 
& 0.3577 & 0.8389 & 0.4953 
& 0.5543 & 0.4457 
& 0.9150 & 0.3287 \\
ESC-Role  
& 0.4373 & 0.7882 & 0.3559 
& 0.7012 & 0.2988 
& 0.8948 & 0.3575 \\
Eeyore    
& 0.5201 & 0.7482 & 0.2687 
& 0.6767 & 0.3233 
& 0.6087 & 0.6454 \\
Ours      
& \textbf{0.5520} & \textbf{0.6901} & \textbf{0.2186} 
& \textbf{0.7433} & \textbf{0.2567} 
& -0.0555 & 0.7351 \\
\bottomrule
\end{tabular}
\caption{Lexical, semantic, and sentiment diversity of seeker utterances across simulators. Arrows indicate whether higher ($\uparrow$) or lower ($\downarrow$) values are better. Sentiment statistics are computed using VADER compound scores.}
\label{tab:seeker_diversity_extended}
\end{table*}



\begin{table*}[t]
\small
\setlength{\tabcolsep}{3.8pt}
\renewcommand{\arraystretch}{1.15}

\begin{tabular}{llrrrrrrrrrr}
\toprule
\multirow{2}{*}{Supporter Model} & \multirow{2}{*}{Seeker Simulator} &
\multicolumn{5}{c}{ES-Skills} & \multicolumn{5}{c}{General-Skills} \\
\cmidrule(lr){3-7}\cmidrule(lr){8-12}
& &
Iden. & Comf. & Sugg. & Expe. & Info. & Cons. & Role. & Expr. & Huma. & Over. \\
\midrule

\multirow{5}{*}{GPT-5-mini}
& ClientCAST
& 4.983 & 4.990 & 3.913 & 2.097 & 3.563 & 5.000 & 5.000 & 5.000 & 5.000 & 5.000 \\
& ESC-Judge
& 4.980 & 4.977 & 4.070 & 2.473 & 3.853 & 5.000 & 5.000 & 5.000 & 5.000 & 5.000 \\
& ESC-Role
& 4.987 & 4.977 & 3.530 & 2.033 & 3.163 & 4.990 & 4.990 & 4.993 & 4.993 & 4.990 \\
& Eeyore
& 4.907 & 4.940 & 3.567 & 2.043 & 3.503 & 5.000 & 5.000 & 4.997 & 4.997 & 4.997 \\
& Ours
& \textbf{4.410} & \textbf{4.477} & \textbf{2.820} & \textbf{1.367} & \textbf{2.393} & \textbf{4.933} & \textbf{4.963} & \textbf{4.860} & \textbf{4.830} & \textbf{4.853} \\
\midrule

\multirow{5}{*}{Llama-3-8B-Instruct}
& ClientCAST
& 4.527 & 4.967 & 4.133 & 2.463 & 4.030 & 4.987 & 4.997 & 4.990 & 4.987 & 4.990 \\
& ESC-Judge
& 4.750 & 4.913 & 4.000 & 2.333 & 3.773 & 5.000 & 5.000 & 5.000 & 5.000 & 5.000 \\
& ESC-Role
& 4.421 & 4.803 & 4.184 & 1.960 & 4.074 & 5.000 & 5.000 & 5.000 & 5.000 & 5.000 \\
& Eeyore
& 4.301 & 4.816 & 3.870 & 1.977 & 3.836 & 5.000 & 5.000 & 4.993 & 4.993 & 4.993 \\
& Ours
& \textbf{4.097} & \textbf{4.297} & \textbf{3.300} & \textbf{1.590} & \textbf{2.970} & \textbf{4.883} & \textbf{4.920} & \textbf{4.860} & \textbf{4.813} & \textbf{4.840} \\
\midrule

\multirow{5}{*}{Llama-CounselChat}
& ClientCAST
& 4.113 & 4.563 & 3.960 & 2.367 & 4.010 & 4.947 & 4.933 & 4.767 & 4.737 & 4.760 \\
& ESC-Judge
& 4.167 & 4.447 & 3.990 & 2.190 & 3.830 & 4.993 & 4.993 & 4.897 & 4.857 & 4.900 \\
& ESC-Role
& 4.053 & 4.227 & 4.313 & 2.057 & 4.313 & 4.990 & 4.983 & 4.850 & 4.793 & 4.853 \\
& Eeyore
& 3.893 & 3.850 & 3.820 & 1.830 & 4.060 & 4.803 & 4.780 & 4.530 & 4.313 & 4.503 \\
& Ours
& \textbf{3.750} & \textbf{3.577} & \textbf{3.530} & \textbf{1.563} & \textbf{3.600} & \textbf{4.560} & \textbf{4.597} & \textbf{4.257} & \textbf{3.960} & \textbf{4.220} \\
\midrule

\multirow{5}{*}{Llama-ESConv}
& ClientCAST
& 3.900 & 4.220 & 3.620 & 2.383 & 3.353 & 4.900 & 4.880 & 4.640 & 4.587 & 4.653 \\
& ESC-Judge
& 4.000 & 4.267 & 3.203 & 2.587 & 2.717 & 4.947 & 4.960 & 4.797 & 4.757 & 4.827 \\
& ESC-Role
& 4.053 & 4.367 & 3.910 & 2.320 & 3.540 & 4.967 & 4.963 & 4.810 & 4.787 & 4.833 \\
& Eeyore
& 3.817 & 3.927 & 3.543 & 2.053 & 3.133 & 4.907 & 4.890 & 4.587 & 4.457 & 4.593 \\
& Ours
& \textbf{3.390} & \textbf{3.150} & \textbf{2.303} & \textbf{1.213} & \textbf{1.807} & \textbf{4.387} & \textbf{4.363} & \textbf{3.960} & \textbf{3.617} & \textbf{3.887} \\
\midrule

\multirow{5}{*}{Llama-ExTES}
& ClientCAST
& 4.427 & 4.897 & 3.993 & 2.257 & 3.997 & 4.993 & 4.993 & 4.983 & 4.973 & 4.980 \\
& ESC-Judge
& 4.523 & 4.870 & 3.797 & 2.267 & 3.753 & 4.997 & 5.000 & 4.997 & 4.997 & 4.997 \\
& ESC-Role
& 4.480 & 4.773 & 4.140 & 2.040 & 4.077 & 5.000 & 5.000 & 5.000 & 5.000 & 5.000 \\
& Eeyore
& 4.333 & 4.660 & 3.720 & 1.903 & 3.797 & 5.000 & 5.000 & 4.990 & 4.980 & 4.990 \\
& Ours
& \textbf{4.083} & \textbf{4.113} & \textbf{3.210} & \textbf{1.580} & \textbf{3.040} & \textbf{4.883} & \textbf{4.923} & \textbf{4.813} & \textbf{4.740} & \textbf{4.810} \\
\midrule

\multirow{5}{*}{Llama-Psych8k}
& ClientCAST
& 4.237 & 4.720 & 3.977 & 2.143 & 3.937 & 4.993 & 4.993 & 4.977 & 4.960 & 4.973 \\
& ESC-Judge
& 4.277 & 4.687 & 4.030 & 2.297 & 3.960 & 5.000 & 5.000 & 4.997 & 4.997 & 4.997 \\
& ESC-Role
& 4.227 & 4.540 & 4.253 & 2.090 & 4.167 & 5.000 & 5.000 & 5.000 & 5.000 & 5.000 \\
& Eeyore
& 4.097 & 4.183 & 3.823 & 1.850 & 3.843 & 4.977 & 4.983 & 4.923 & 4.893 & 4.927 \\
& Ours
& \textbf{4.000} & \textbf{3.803} & \textbf{3.573} & \textbf{1.643} & \textbf{3.517} & \textbf{4.867} & \textbf{4.883} & \textbf{4.733} & \textbf{4.587} & \textbf{4.710} \\
\midrule

\multirow{5}{*}{Llama-PsyInsight}
& ClientCAST
& 4.070 & 4.243 & 3.617 & 1.743 & 2.847 & 4.970 & 4.967 & 4.900 & 4.870 & 4.900 \\
& ESC-Judge
& 4.297 & 4.323 & 3.053 & 1.750 & 2.543 & 4.970 & 4.993 & 4.890 & 4.877 & 4.917 \\
& ESC-Role
& 4.167 & 4.493 & 3.897 & 1.940 & 3.443 & 5.000 & 5.000 & 4.977 & 4.963 & 4.977 \\
& Eeyore
& 3.837 & 3.563 & 3.040 & 1.433 & 2.537 & 4.870 & 4.790 & 4.653 & 4.457 & 4.623 \\
& Ours
& \textbf{3.503} & \textbf{3.017} & \textbf{1.983} & \textbf{1.087} & \textbf{1.547} & \textbf{4.418} & \textbf{4.431} & \textbf{4.167} & \textbf{3.843} & \textbf{4.070} \\
\bottomrule
\end{tabular}

\caption{Merged results across supporter models and seeker simulators. Scores are averaged and rounded to three decimals. Within each supporter, the lowest score per metric is boldfaced.}
\label{tab:merged_metrics_10cols}
\end{table*}

\begin{figure*}[t]
  \centering

  \begin{subfigure}[t]{0.49\textwidth}
    \centering
    \includegraphics[width=\textwidth]{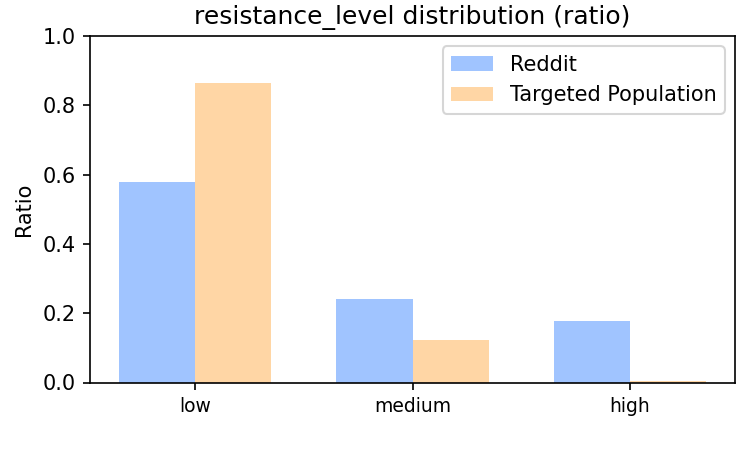}
    \caption{Resistance level distribution}
    \label{fig:resistance-compare}
  \end{subfigure}
  \hfill
  \begin{subfigure}[t]{0.49\textwidth}
    \centering
    \includegraphics[width=\textwidth]{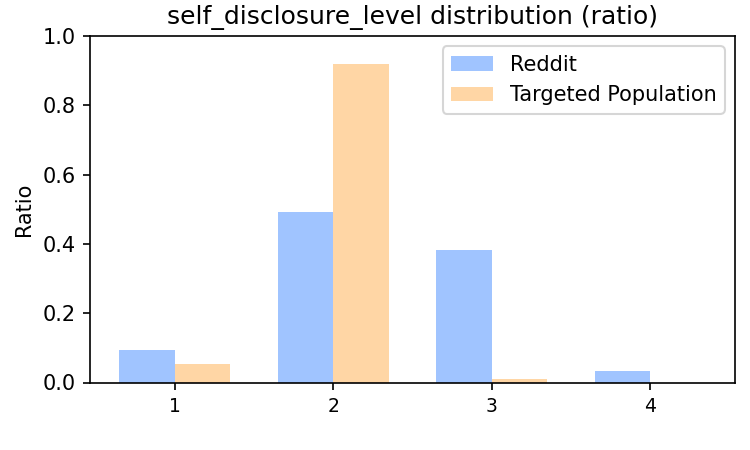}
    \caption{Self-disclosure level distribution}
    \label{fig:selfdisclosure-compare}
  \end{subfigure}

  \vspace{12pt}

  \begin{subfigure}[t]{0.49\textwidth}
    \centering
    \includegraphics[width=\textwidth]{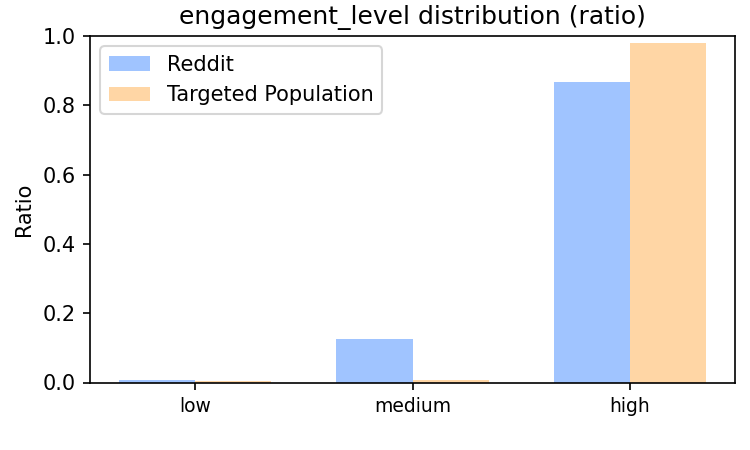}
    \caption{Engagement level distribution}
    \label{fig:engagement-compare}
  \end{subfigure}
  \hfill
  \begin{subfigure}[t]{0.49\textwidth}
    \centering
    \includegraphics[width=\textwidth]{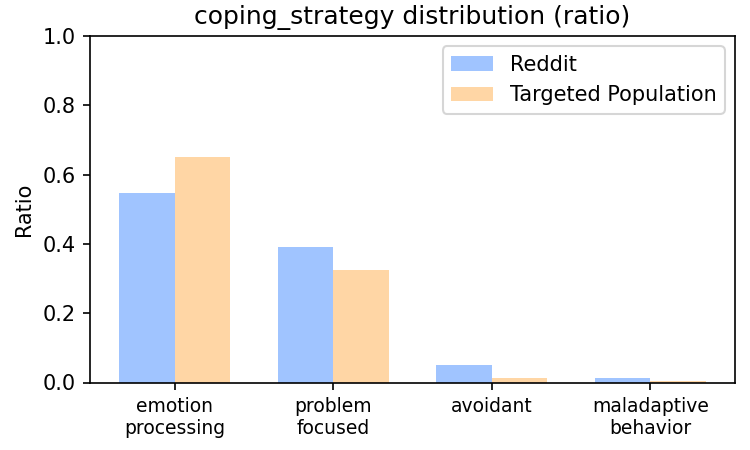}
    \caption{Coping strategy distribution}
    \label{fig:coping-compare}
  \end{subfigure}

  \vspace{12pt}

  \begin{subfigure}[t]{0.49\textwidth}
    \centering
    \includegraphics[width=\textwidth]{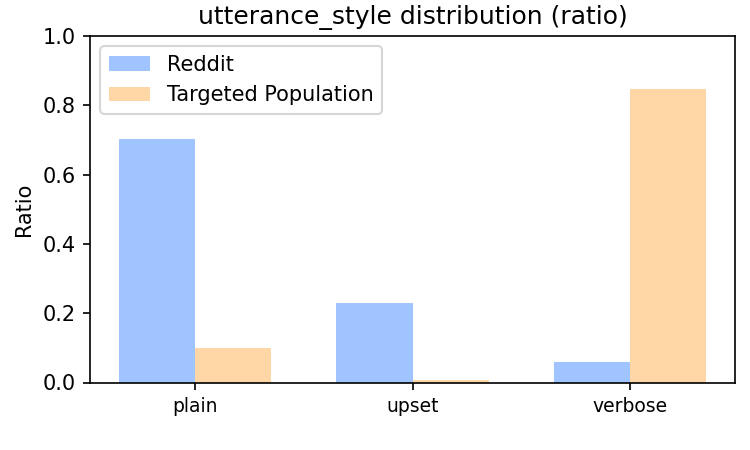}
    \caption{Utterance style distribution}
    \label{fig:utterance-compare}
  \end{subfigure}
  \hfill
  \begin{subfigure}[t]{0.49\textwidth}
    \centering
    \includegraphics[width=\textwidth]{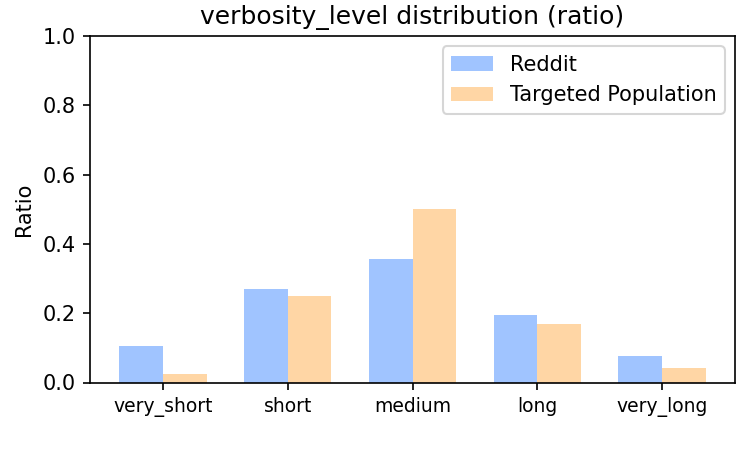}
    \caption{Verbosity level distribution}
    \label{fig:verbosity-compare}
  \end{subfigure}

  \vspace{12pt}

  \begin{subfigure}[t]{0.49\textwidth}
    \centering
    \includegraphics[width=\textwidth]{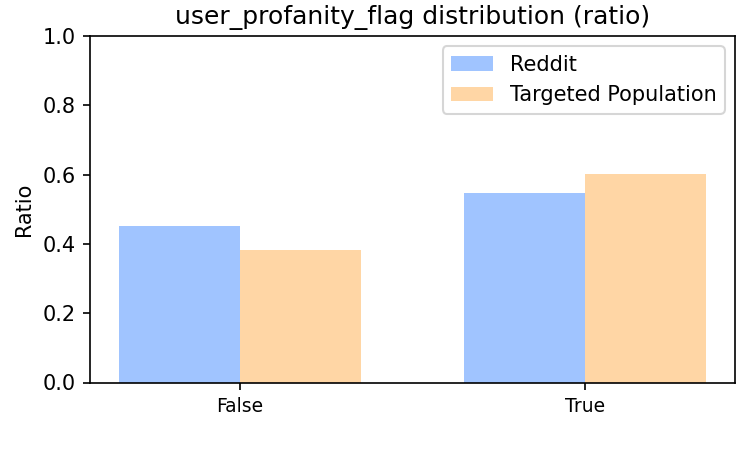}
    \caption{Profanity flag distribution}
    \label{fig:profanity-compare}
  \end{subfigure}
  \hfill
  \begin{subfigure}[t]{0.49\textwidth}
    \centering
    \includegraphics[width=\textwidth]{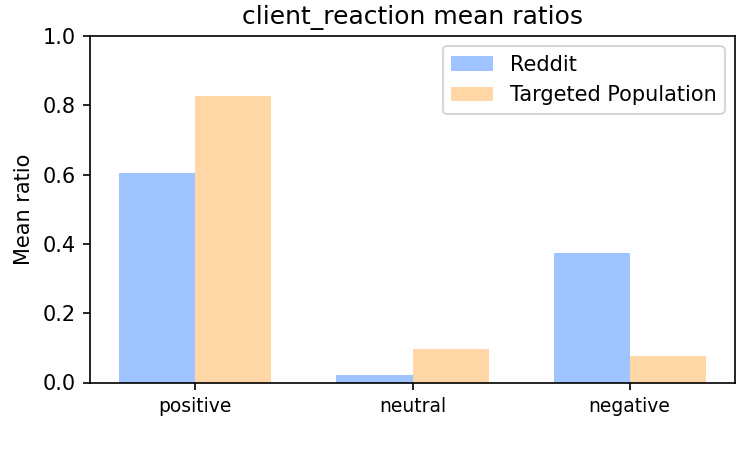}
    \caption{Client reaction ratio (means)}
    \label{fig:reaction-compare}
  \end{subfigure}

  \caption{Comparison of seeker feature distributions between a target seeker population and our Reddit-based training data.}
  \label{fig:realworld-feature-compare-all}
\end{figure*}

\end{document}